\documentclass[10pt,twocolumn,letterpaper]{article}

\usepackage[pagenumbers]{cvpr} 

\usepackage[table,x11names]{xcolor}
\usepackage{soul}
\usepackage{microtype}
\usepackage{xspace}
\usepackage{pifont}
\definecolor{realtimecolor}{rgb}{1.,1,0.85}
\definecolor{CheckGreen}{rgb}{0, 0.55, 0}
\definecolor{XRed}{RGB}{180,0,0}
\usepackage[accsupp]{axessibility}

\makeatletter
\renewcommand\paragraph{\@startsection{paragraph}{4}{\z@}%
	{0.75ex \@plus.5ex \@minus.2ex}%
	{-1em}%
	{\normalfont\normalsize\bfseries\maybe@addperiod}}
\newcommand{\maybe@addperiod}[1]{#1\@addpunct{.}}
\makeatother

\newcommand{\method}{HybridNeRF\xspace}

\usepackage{mathtools}

\newcommand{\norm}[1]{\left\lVert#1\right\rVert}
\DeclarePairedDelimiter{\ceil}{\lceil}{\rceil}
\DeclarePairedDelimiter{\floor}{\lfloor}{\rfloor}
\newcommand{\vct}[1]{\boldsymbol{#1}}
\newcommand{\xmark}{{\color{XRed}\ding{55}}}

\definecolor{cvprblue}{rgb}{0.21,0.49,0.74}
\usepackage[pagebackref,breaklinks,colorlinks,citecolor=cvprblue]{hyperref}
\usepackage{mathtools}

\def\paperID{}
\def\confName{}
\def\confYear{}

\title{\method: Efficient Neural Rendering via Adaptive Volumetric Surfaces}

\author{
	Haithem Turki\textsuperscript{1, 2} \quad
	Vasu Agrawal\textsuperscript{1} \quad
	Samuel Rota Bulò\textsuperscript{1} \quad
	Lorenzo Porzi\textsuperscript{1} \\
	Peter Kontschieder\textsuperscript{1} \quad
	Deva Ramanan\textsuperscript{2} \quad
	Michael Zollhöfer\textsuperscript{1} \quad
	Christian Richardt\textsuperscript{1} \\[0.5em]
	$^1$Meta Reality Labs \quad
	$^2$Carnegie Mellon University
}

\begin{document}
\maketitle

\begin{abstract}
Neural radiance fields provide state-of-the-art view synthesis quality but tend to be slow to render.
One reason is that they make use of volume rendering, thus requiring many samples (and model queries) per ray at render time.
Although this representation is flexible and easy to optimize, most real-world objects can be modeled more efficiently with surfaces instead of volumes, requiring far fewer samples per ray.
This observation has spurred considerable progress in surface representations, such as signed distance functions, but these may struggle to model semi-opaque and thin structures.
We propose a method, \method, that leverages the strengths of both representations by rendering most objects as surfaces while modeling the (typically) small fraction of challenging regions volumetrically.
We evaluate \method against the challenging Eyeful Tower dataset \cite{XuALGBKRPKBLZR2023} along with other commonly used view synthesis datasets.
When comparing to state-of-the-art baselines, including recent rasterization-based approaches,
we improve error rates by 15--30\% while achieving real-time framerates (at least 36 FPS) for virtual-reality resolutions (2K$\times$2K).
Project page:
\href{https://haithemturki.com/hybrid-nerf/}{https://haithemturki.com/hybrid-nerf/}.
\end{abstract}

\section{Introduction}

\begin{figure}
    \captionsetup[subfigure]{justification=centering}
        \centering
        \begin{subfigure}[b]{0.49\linewidth}
                    \caption*{\textbf{RGB}}
            \centering
            \includegraphics[width=\textwidth,trim={0cm 0cm 0cm 0cm},clip]{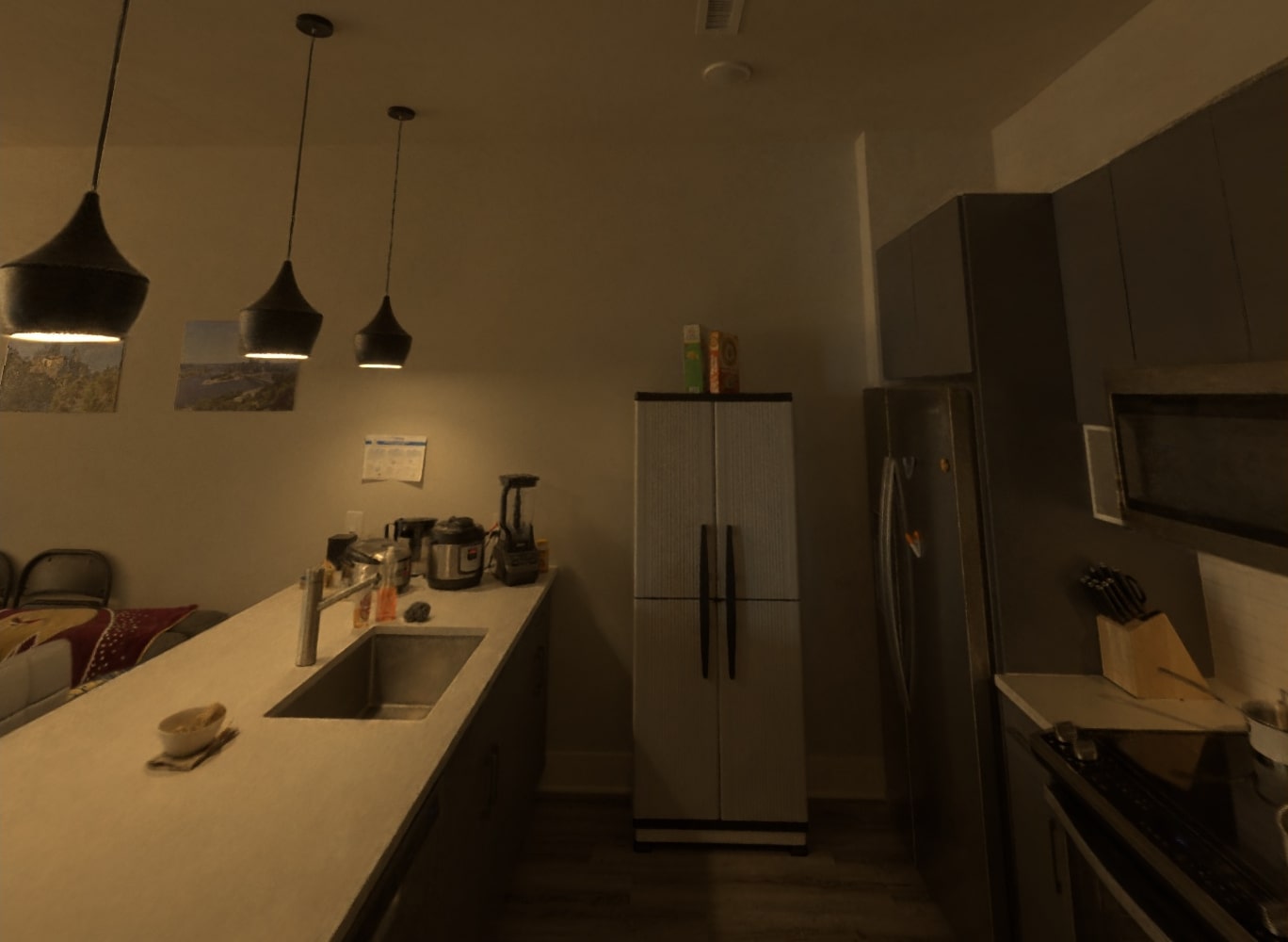}
        \end{subfigure}
        \hfill
        \begin{subfigure}[b]{0.49\linewidth}  
            \centering 
                    \caption*{\textbf{Surfaceness}}
            \includegraphics[width=\textwidth,trim={0cm 0cm 0cm 0cm},clip]{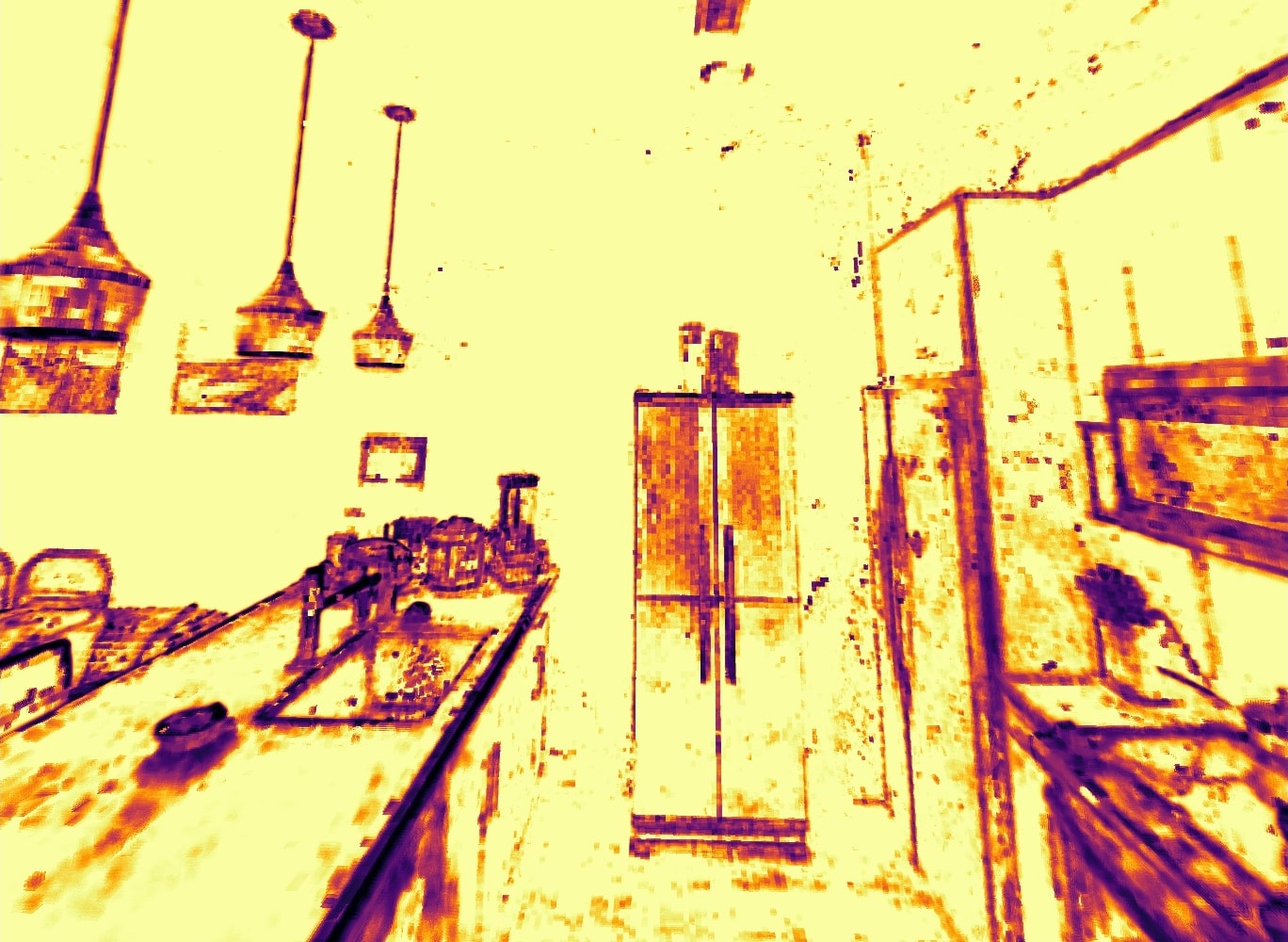}
        \end{subfigure}
        \smallskip
        \begin{subfigure}[b]{0.49\linewidth}  
            \centering 
                    \caption*{\textbf{NeRF\\($\approx$40 samples / ray)}}
            \includegraphics[width=\textwidth,trim={0cm 0cm 0cm 0cm},clip]{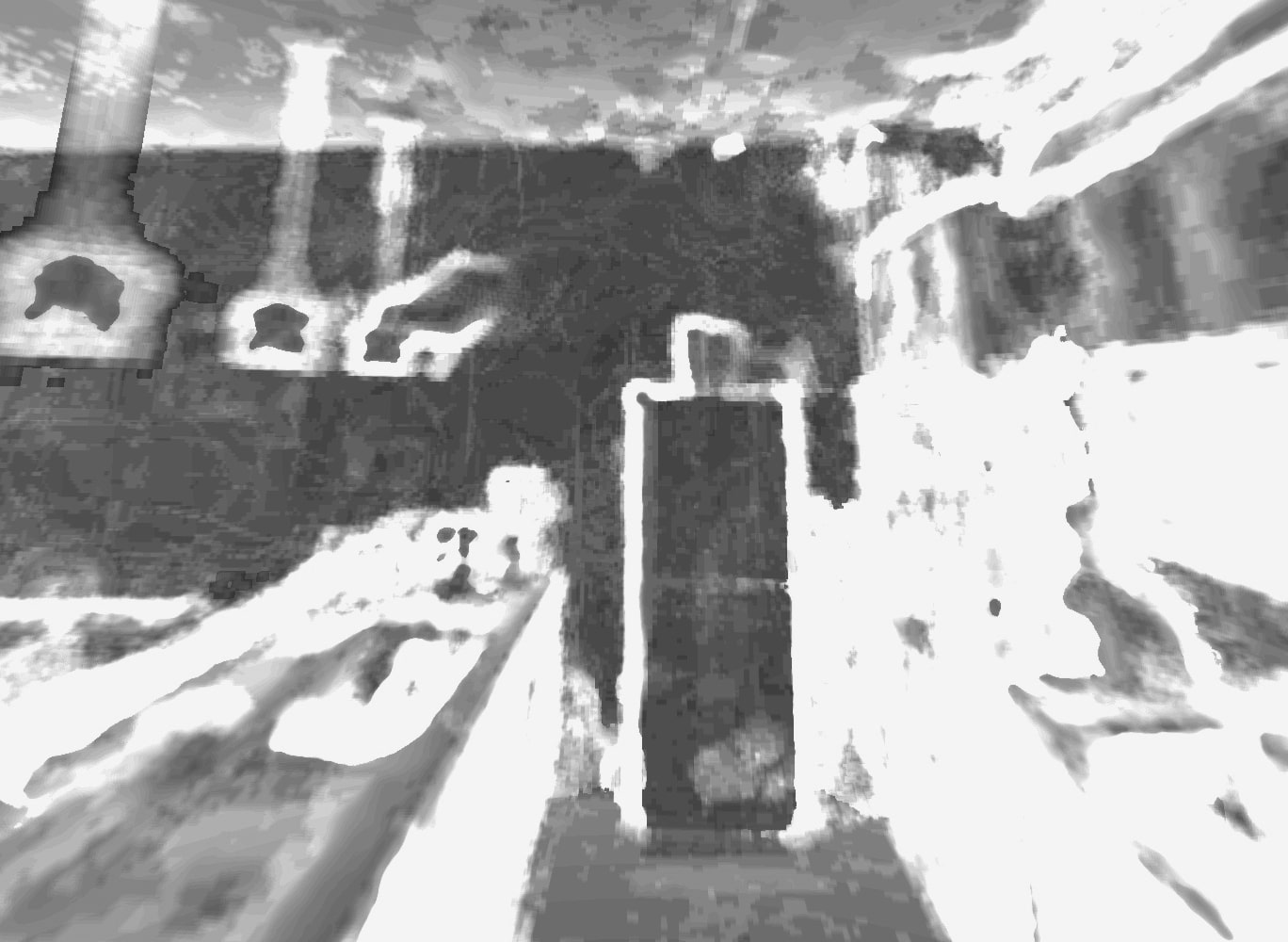}
        \end{subfigure}
        \begin{subfigure}[b]{0.49\linewidth}  
            \centering 
                    \caption*{\textbf{\method\\($\approx$8 samples / ray)}}
            \includegraphics[width=\textwidth,trim={0cm 0cm 0cm 0cm},clip]{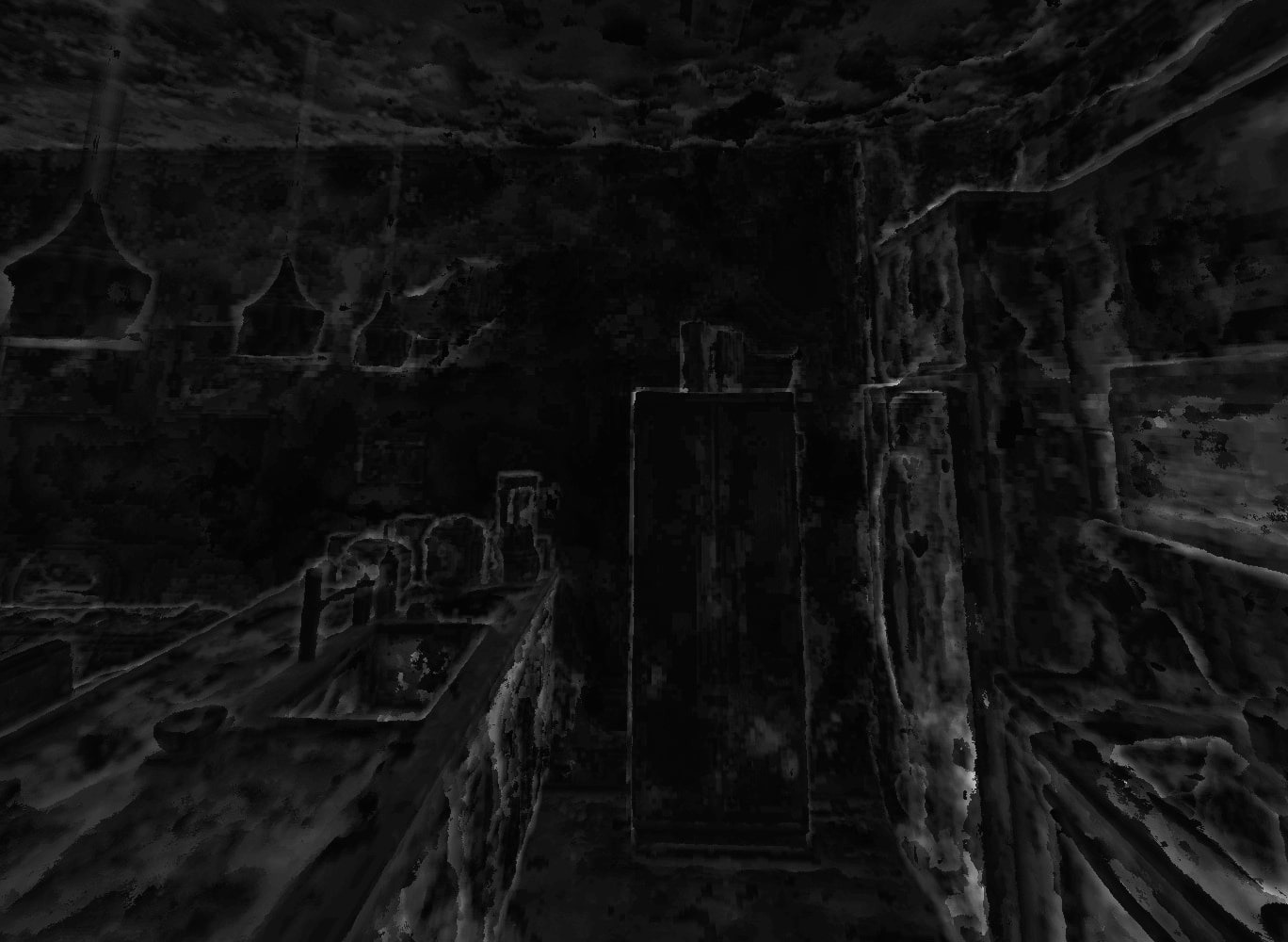}
        \end{subfigure}
 \caption{\textbf{\method.} 
    We train a hybrid surface--volume representation via \emph{surfaceness} parameters that allow us to render most of the scene with few samples.
    We track Eikonal loss as we increase surfaceness to avoid degrading quality near fine and translucent structures (such as wires).
    On the bottom, we visualize the number of samples per ray (brighter is higher).
    Our model renders in high fidelity at 2K$\times$2K resolution at real-time frame rates.}
\vspace{-4mm}
\label{fig:teaser}
\end{figure}

Recent advances in volumetric rendering of neural radiance fields \cite{mildenhall2020nerf} (NeRFs) have led to significant progress towards photorealistic novel-view synthesis.
However, while NeRFs provide state-of-the-art rendering quality, they remain slow to render.

\paragraph{Efficiency}

We seek to construct a representation that enables high-quality efficient rendering, which is necessary for immersive applications, such as augmented reality or virtual teleconferencing.

While recent rasterization-based techniques, such as mesh baking \cite{YarivHRVSSBM2023,ChenXGYS2022} or Gaussian splatting \cite{KerblKLD2023}, are very efficient, they still struggle to capture transparent or fine structures, and view-dependent effects (like reflections or specularities), respectively.
Instead, we focus on NeRF's standard ray casting paradigm, and propose techniques that enable a better speed--quality trade-off.

\paragraph{Rendering}

We start with the observation that neural implicit surface representations, such as signed distance functions (SDFs), which were originally proposed to improve the geometry quality of NeRFs via regularization \cite{YarivGKL2021, WangLLTKW2021}, can \emph{also} be used to dramatically increase efficiency by requiring fewer samples per ray.
In the limit, only a single sample on the surface is required.
In practice, renderers still need to identify the location of the target sample(s), which can be done by generating samples via an initial proposal network \cite{BarroMVSH2022} or other techniques, such as sphere tracing \cite{liu2020dist}.

\begin{figure*}
    \includegraphics[width=\textwidth]{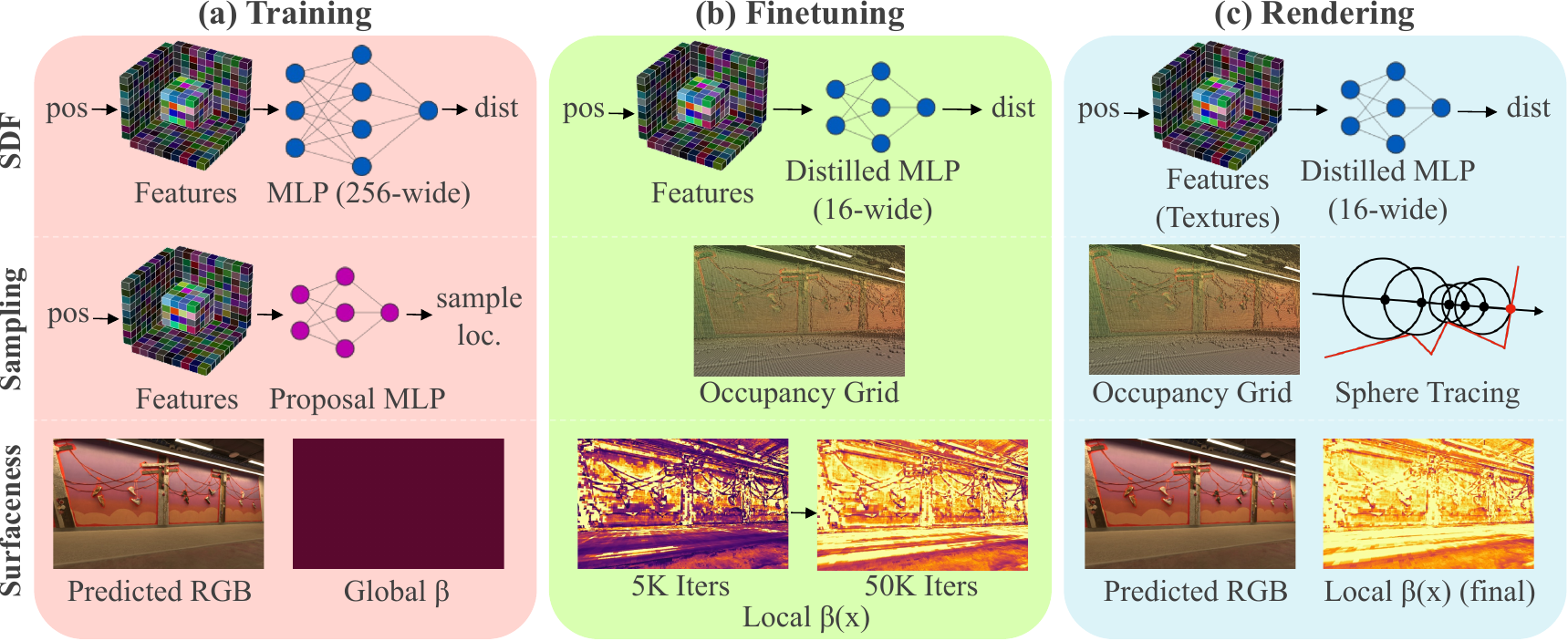}
	\caption{\textbf{Approach.}
        In the first phase of our pipeline (\textbf{a}), we train a VolSDF-like \cite{YarivGKL2021} model with distance-adjusted Eikonal loss to model backgrounds without a separate NeRF (\cref{sec:backgrounds}).
        We then crucially transition from a uniform surfaceness parameter $\beta$ to position-dependent $\beta(\vct x)$ values to model most of the scene as thin surfaces (needing few samples) without degrading quality near fine and semi-opaque structures (\textbf{b}).
        Since our model behaves as a valid SDF in $>$95\% of the scene, we use sphere tracing at render time (\textbf{c}) along with lower-level optimizations (hardware texture interpolation) to query each sample as efficiently as possible.
    }
    \label{fig:approach}
 \vspace*{-4mm}
\end{figure*}

\paragraph{Surfaceness}

While surface-based neural fields are convenient for rendering, they often struggle to reconstruct scenes with thin structures or view-dependent effects, such as reflections and translucency.
This is one reason that surfaces are often transformed into volumetric models for rendering~\cite{YarivGKL2021}.
A crucial transformation parameter is a scalar temperature $\beta$ that is used to convert a $\beta$-scaled signed distance value into a density.
Higher temperatures tend to produce an ideal binary occupancy field that can improve rendering speed but can struggle for challenging regions as explained above.
Lower temperatures allow the final occupancy field to remain flexible, whereby the $\beta$-scaled SDF essentially acts as a reparameterization of the underlying occupancy field.
As such, we refer to $\beta$ as the {\em surfaceness} of the underlying scene (see \cref{fig:teaser}).
Prior work treats $\beta$ as a global parameter that is explicitly scheduled or learned via gradient descent~\cite{YarivGKL2021}.
We learn it in a spatially adaptive manner.

\paragraph{Contributions}
Our primary contribution is 
a hybrid surface--volume representation that combines the best of both worlds.
Our key insight is to replace the global parameter $\beta$ with spatially-varying parameters $\beta(\vct{x})$ corresponding to the surfaceness of regions in the 3D scene.
At convergence, we find that most of the scene ($>95\%$) can be efficiently modeled as a surface.
This allows us to render with far fewer samples than fully volumetric methods, while achieving higher fidelity than pure surface-based approaches.
Additionally,
\begin{enumerate}
    \item
    We propose a weighted Eikonal regularization that allows our method to render high-quality complex backgrounds without a separate background model.
    \item
    We implement specific rendering optimizations, such as hardware texture interpolation and sphere tracing, to significantly accelerate rendering at high resolutions.

    \item
    We present state-of-the-art reconstruction results on three different datasets, including the challenging Eyeful Tower dataset \cite{XuALGBKRPKBLZR2023}, while rendering almost 10$\times$ faster.
\end{enumerate}

\section{Related Work}

Many works try to accelerate the rendering speed of neural radiance fields (NeRF).
We discuss a representative selection of such approaches below.

\paragraph{Voxel baking}

Some of the earliest NeRF acceleration methods store precomputed non-view dependent model outputs, such as spherical harmonics coefficients, into finite-resolution structures \cite{yu2021plenoctrees, hedman2021snerg, GarbiKJSV2021, ChenFHT2023, DuckwHRZTLSB2023}.
These outputs are combined with viewing direction to compute the final radiance at render time, bypassing the original model entirely.
Although these methods render extremely quickly (some $>$200 FPS \cite{GarbiKJSV2021}),
they are limited by the finite capacity of the caching structure and cannot capture fine details at room scale.

\paragraph{Feature grids}

Recent methods use a hybrid approach that combines a learned feature grid with a much smaller MLP than the original NeRF \cite{ChenXGYS2022, FridoMWRK2023, MuelleESK2022}.
Instant-NGP \cite{MuelleESK2022} (iNGP), arguably the most popular of these methods, encodes features into a multi-resolution hash table.
Although these representations speed up rendering, they cannot reach the level needed for real-time HD rendering alone, as even iNGP reaches less than 10 FPS on real-world datasets at high resolution.
MERF \cite{ReiseSVSMGBH2023} comes closest through a baking pipeline that uses various sampling and memory layout optimizations that we also make use of in our implementation.

\paragraph{Surface--volume representations}

Several methods \cite{OechsPG2021, WangLLTKW2021, YarivGKL2021} derive density values from the outputs of a signed distance function, which are then rendered volumetrically as in NeRF.
These hybrid representations retain NeRF's ease of optimization while improving surface geometry.
Follow-ups \cite{YarivHRVSSBM2023, GuoCWHSQZ2023} bake the resulting surface geometry into a mesh that is further optimized and simplified.
Similar to early voxel-baking approaches, these methods render quickly ($>$70 FPS) but are limited by the capacity of the mesh and texture, and thus struggle to model thin structures, transparency, and view-dependent effects.
We train a similar SDF representation in our method but continue using the base neural model at render time.
Concurrent to our work, Adaptive shells \cite{WangSNSGKFMG2023} augments NeuS \cite{WangLLTKW2021} with a spatially-varying kernel similar to our adaptive surfaceness described in \cref{sec:finetuning}.

\paragraph{Sample efficiency}

Several approaches accelerate rendering by intelligently placing far fewer samples along each ray than the original hierarchical strategy proposed by NeRF \cite{neff2021donerf, PialaC2021, KurzNLZS2022, BarroMVSH2022, GuptaHXLSSCB2023}.
These methods all train auxiliary networks that are cheaper to evaluate than the base model.
However, as they are based on purely volumetric representations, they are limited in practice as to how few samples they can use per ray without degrading quality, and therefore exhibit a different quality--performance tradeoff curve than ours.

\paragraph{Gaussians}
Recent methods take inspiration from NeRF's volume rendering formula but discard the neural network entirely and instead parameterize the scene through a set of 3D Gaussians \cite{KerblKLD2023, keselman2022fuzzy, keselman2023fuzzyplus, wang2022voge}.
Of these, 3D Gaussian splatting \cite{KerblKLD2023} has emerged as the new state of the art, rendering at $>$100 FPS with higher fidelity than previous non-neural approaches.
Although encouraging, it is sensitive to initialization (especially in far-field areas) and limited in its ability to reason about inconsistencies within the training dataset (such as transient shadows) and view dependent effects.

\section{Method}

Given a collection of RGB images and camera poses, our goal is to learn a 3D representation that generates novel views at VR resolution (at least 2K$\times$2K pixels) in real-time (at least 36 FPS),
while achieving a high degree of visual fidelity.
As we target captures taken under real-world conditions, our representation must be able to account for inconsistencies across training images due to lighting changes and shadows (even in ``static" scenes).
We build upon NeRF's raycasting paradigm, which can generate highly photorealistic renderings, and improve upon its efficiency.
As the world mostly consists of surfaces, we train a representation that can render surfaces with few samples and without degrading the rest of the scene.
We outline our method in \cref{fig:approach} and
present our model architecture and the first training stage in \cref{sec:representation}, which is followed by finetuning of our model to accelerate rendering without compromising quality in \cref{sec:finetuning}.
We discuss how to model unbounded scenes in \cref{sec:backgrounds} and present final render-time optimizations in \cref{sec:rendering}.

\subsection{Representation}
\label{sec:representation}

\paragraph{Preliminaries}
NeRF \cite{mildenhall2020nerf} represents a scene as a continuous volumetric radiance field that encodes the scene's geometry and view-dependent appearance within the weights of an MLP.
NeRF renders pixels by sampling positions $\vct x_i$ along the corresponding camera ray, querying the MLP to obtain density and color values, $\sigma_i\coloneqq\sigma(\vct x_i)$ and $\mathbf{c}_i \coloneqq \vct c(\vct x_i, \vct d_r)$, respectively (with $\vct d_r$ as the ray direction).
The density values $\sigma_i$ are converted into opacity values $\alpha_i \coloneqq 1 - \exp(-\sigma_i \delta_i)$, where $\delta_i$ is the distance between samples.
The final ray color $\hat{\vct{c}}_r\coloneqq\sum_{i=0}^{N-1}\vct c_iw_i$ is obtained as the combination of the color samples $\vct c_i$ with weights $w_i\coloneqq\exp( -\!\sum_{j=0}^{i-1} \!\sigma_j \delta_j) \alpha_i$.
The training process optimizes the model by sampling batches of image pixels and minimizing the L2 reconstruction loss.
We refer to \citet{mildenhall2020nerf} for details.

\begin{figure}
\captionsetup[subfigure]{justification=centering}
        \centering
        \begin{subfigure}[b]{0.49\linewidth}
            \caption*{\textbf{NeRF ($\approx$35 samples / ray)}}
        \centering
        \includegraphics[width=\textwidth]{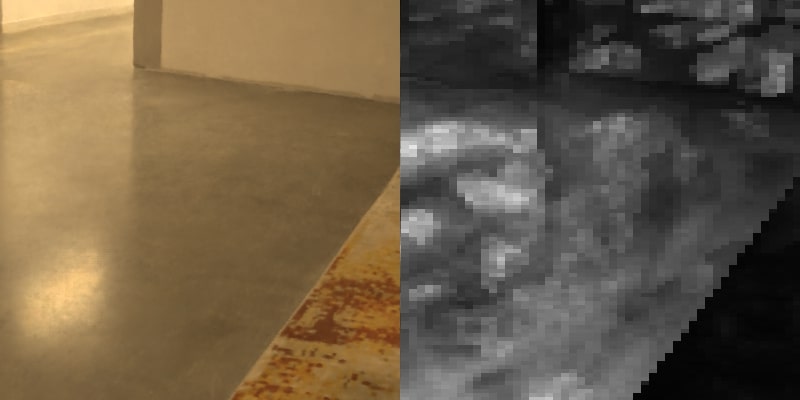}
        \end{subfigure}
        \hfill
        \begin{subfigure}[b]{0.49\linewidth}  
            \centering 
             \caption*{\textbf{\method ($\approx$9 samples / ray)}}
            \includegraphics[width=\textwidth]{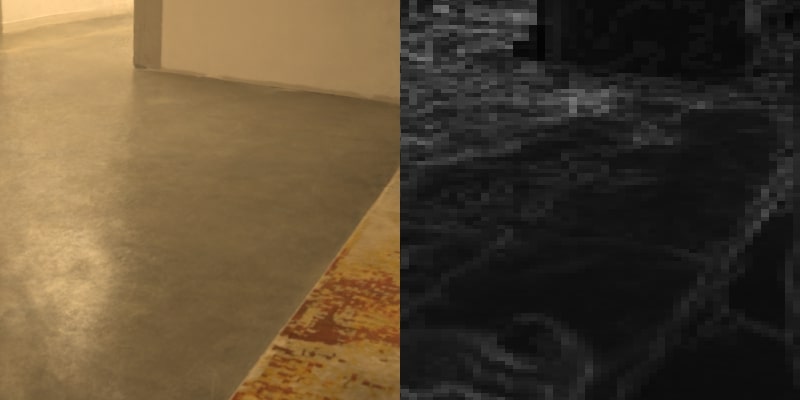}
        \end{subfigure}
        \smallskip
        \begin{subfigure}[b]{0.49\linewidth}
            \centering
            \includegraphics[width=\textwidth]{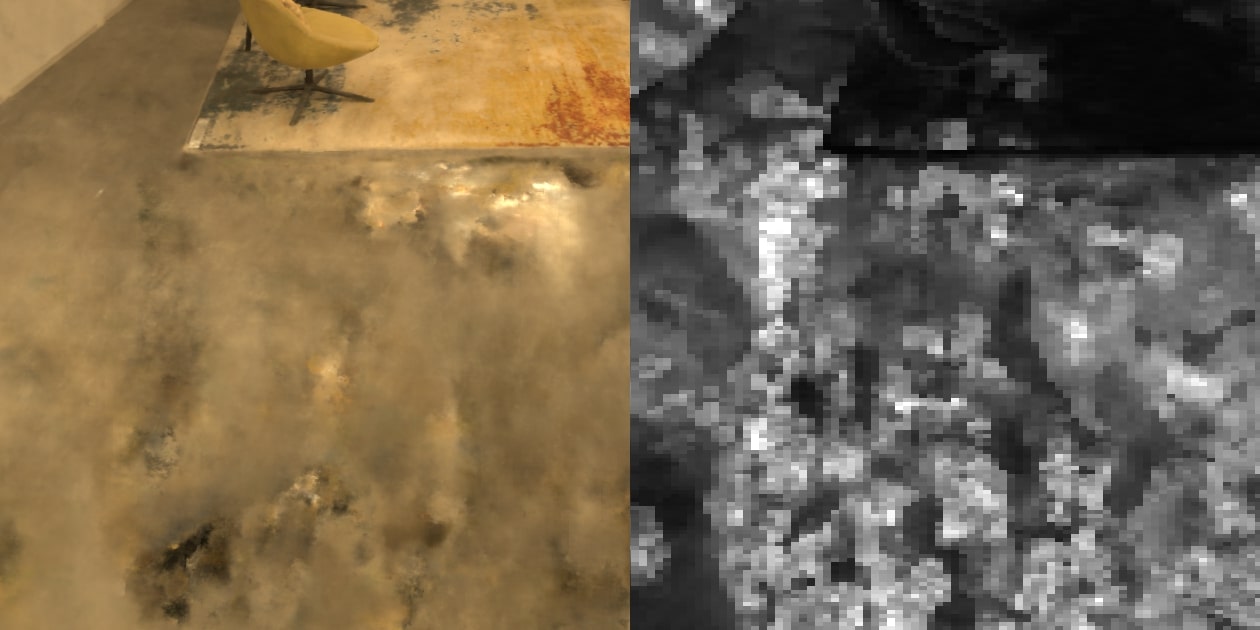}
        \end{subfigure}
        \hfill
        \begin{subfigure}[b]{0.49\linewidth}  
            \centering 
            \includegraphics[width=\textwidth]{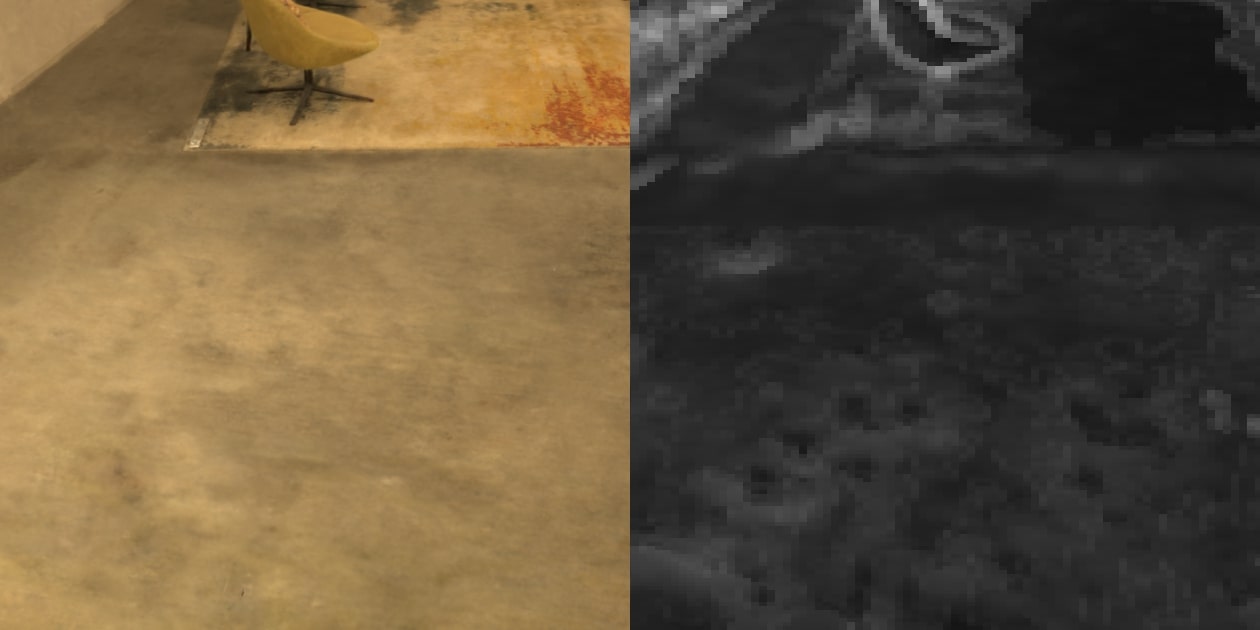}
        \end{subfigure}
	\caption{\textbf{Surfaces.}
		Since NeRF directly predicts density, it often `cheats' by modeling specular surfaces, such as floors, as semi-transparent volumes that require many samples per ray (heatmaps shown on the \textbf{right}, with brighter values corresponding to more samples).
		Methods that derive density from signed distances, such as ours, improve surface geometry and appearance while using fewer samples per ray.
	}
 \vspace*{-4mm}
	\label{fig:apparent-surfaces}
\end{figure}

\paragraph{Modeling density}
The original NeRF representation has the flexibility of representing semi-transparent surfaces, for the density field is not forced to
saturate.
However, the model often abuses this property
by generating semi-transparent volumes to mimic reflections and other view-dependent effects (\cref{fig:apparent-surfaces}).
This hampers our goal of minimizing the samples per ray needed for rendering.

To address this problem, surface--volume representations \cite{OechsPG2021, YarivGKL2021, WangLLTKW2021} learn well-defined surfaces by interpreting MLP outputs $f(\vct x)$ as a signed distance field (SDF) to represent scene surfaces as the zero-level set of the function $f$.

As the norm of the gradient of an SDF should typically be 1, the MLP is regularized via the Eikonal loss:
\begin{align}
    \mathcal{L}_{\text{Eik}}(\mathbf{r}) \coloneqq \sum_{i=0}^{N-1} \eta_i(\norm{\nabla f(\vct x_i)}-1)^2,
    \label{eq:eikonal-loss1}
\end{align}
where $\eta_i$ is a per-sample loss weight typically set to $1$.
The signed distances are converted into densities $\sigma_\text{SDF}$ that are paired with color predictions, and rendered as in NeRF.
Specifically, we follow VolSDF's approach \cite{YarivGKL2021} and define:
\begin{equation}
    \sigma_\text{SDF}(\vct{x}) \coloneqq \beta(\vct x) \Psi(f(\vct{x})\beta(\vct x)) \text{,}
    \label{eq:volsdf-density}
\end{equation}
where $\beta(\vct x)>0$ determines the \emph{surfaceness} of point $\vct x$, \ie how concentrated the density should be around the zero-level set of $f$, and $\Psi$ is the CDF of a standard Laplace distribution:
\vspace{-\baselineskip} 
\begin{align}
	\Psi(s) = \begin{cases}
		\frac{1}{2}     \exp(-s) & \text{if } s > 0  \\
		1 - \frac{1}{2} \exp(s)  & \text{if } s\leq 0 \text{.}
	\end{cases}
	\label{eq:sdf-density}
\end{align}
In prior works, the surfaceness $\beta(\vct x)$ is independent of position $\vct x$.
We instead consider a surfaceness \emph{field} implemented as a $512^3$ grid of values queried via nearest-neighbor interpolation.
We first constrain the surfaceness parameters to be globally uniform, and allow them to diverge spatially during the finetuning stage (\cref{sec:finetuning}).

\begin{figure}
    \captionsetup[subfigure]{justification=centering}
        \centering
        \begin{subfigure}[b]{0.32\linewidth}
                    \caption*{\textbf{Global $\beta(x)=100$ \\ ($\approx$30 samples / ray)}}
            \centering
            \includegraphics[width=\textwidth]{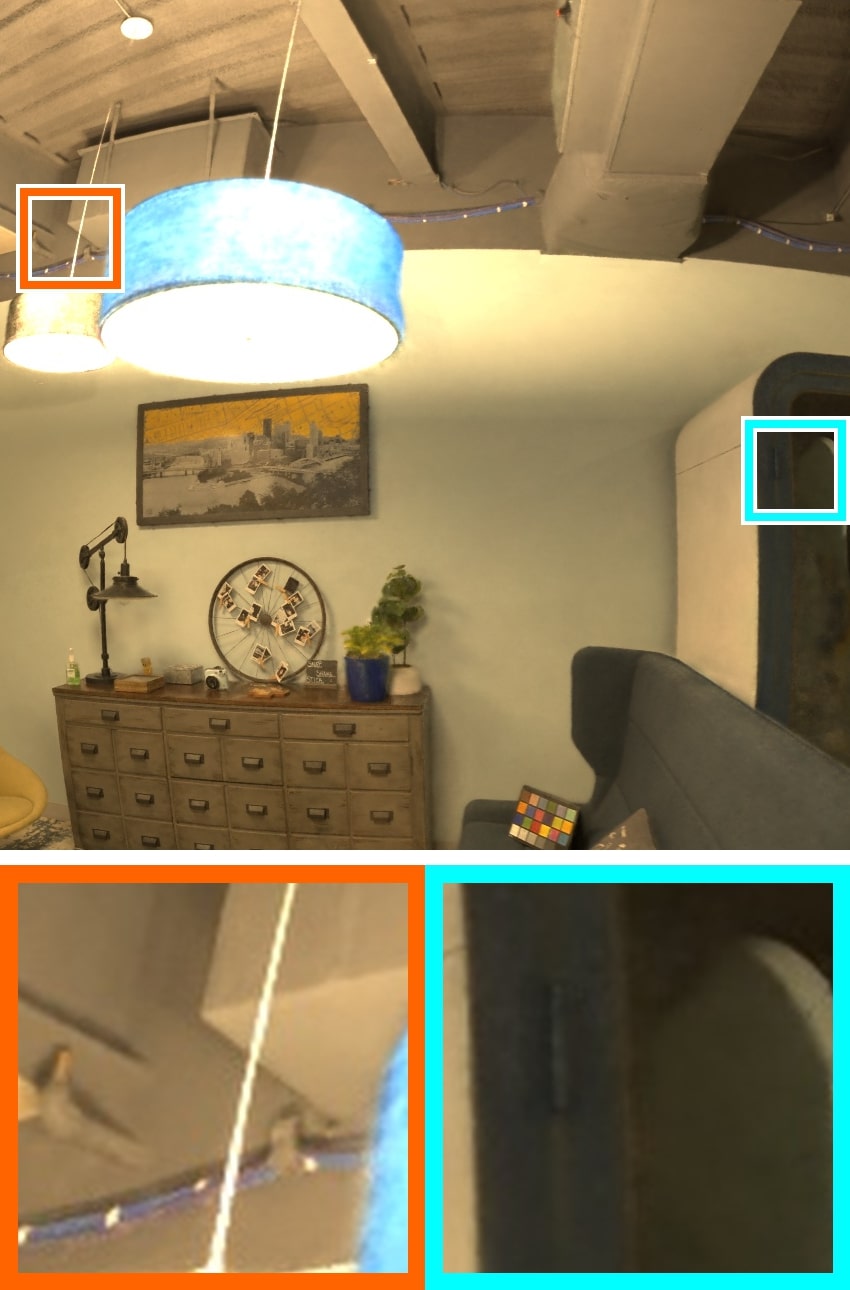}
        \end{subfigure}
        \hfill
        \begin{subfigure}[b]{0.32\linewidth}  
            \centering 
             \caption*{\textbf{Global $\beta(x)=2000$ \\ ($\approx$6 samples / ray)}}
            \includegraphics[width=\textwidth]{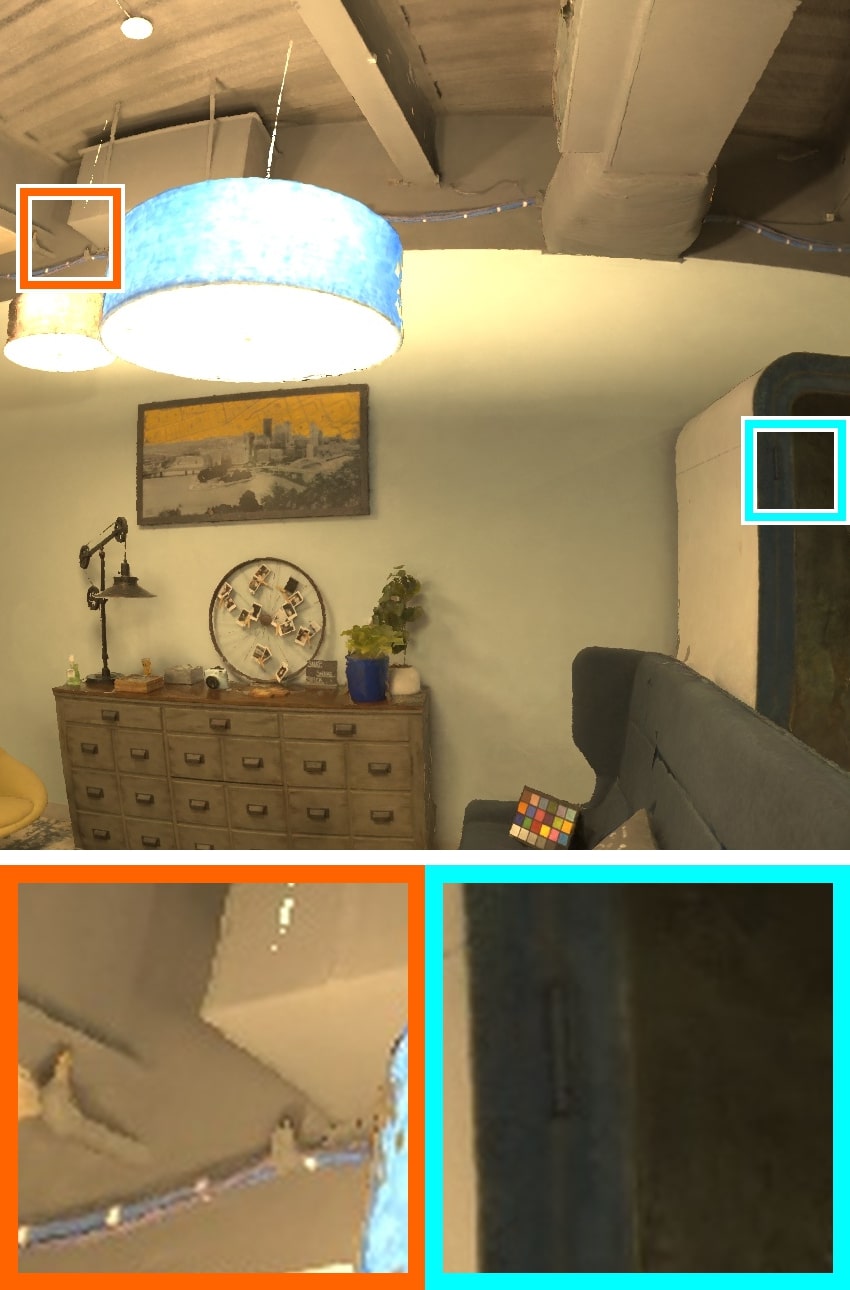}
        \end{subfigure}
        \hfill
        \begin{subfigure}[b]{0.32\linewidth}   
            \centering 
            \caption*{\textbf{Adaptive $\beta(x)$ \\ ($\approx$8 samples / ray)}}
            \includegraphics[width=\textwidth]{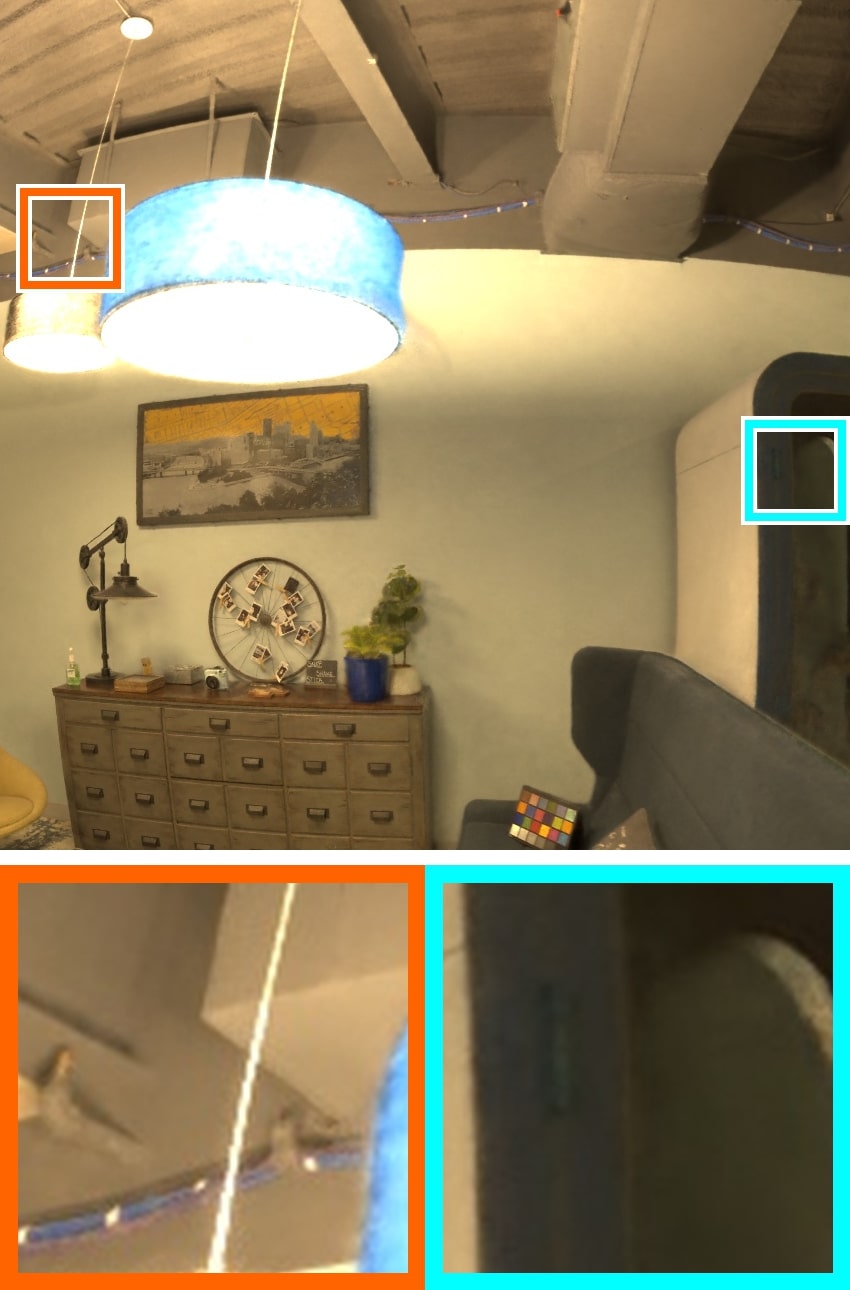}
        \end{subfigure}	
	\caption{\textbf{Choice of $\beta$.}
		Increasing $\beta$ reduces the number of samples needed to render per ray, but negatively impacts quality near fine objects (lamp wires) and transparent structures (glass door).
	}
\vspace*{-4mm}
	\label{fig:different-beta}
\end{figure}

\paragraph{Model architecture}

We use dense multi-resolution 3D feature grids in combination with multi-resolution triplanes \cite{CaoJ2023,FridoMWRK2023} to featurize 3D sample locations.
We predict color $\vct{c}$ and signed distance $f$ with separate grids, each followed by an MLP,
and use a small
proposal network similar to that used by Nerfacto \cite{TanciWNLYWKASAMKK2023} to improve sampling efficiency.
For a given 3D point,
we fetch $K \!=\! 4$ features per level from (1) the 3D feature grids at 3 resolution levels (128\textsuperscript{3}, 256\textsuperscript{3} and 512\textsuperscript{3})
via trilinear interpolation,
and (2) from triplanes at 7 levels (from 128\textsuperscript{2} to 8,192\textsuperscript{2}) via bilinear interpolation.
We sum the features across levels (instead of concatenation \cite{MuelleESK2022, FridoMWRK2023}),
and concatenate the summed features from the 3D grid to those from the 3 triplanes to obtain a $4K \!=\! 16$-dimensional MLP input.
We encode viewing direction through spherical harmonics (up to the 4th degree) as an auxiliary input to the color MLP.
As our feature grid is multi-resolution, we handle aliasing as in VR-NeRF \cite{XuALGBKRPKBLZR2023}.
See \cref{sec:lod} and \cref{sec:model-architecture} for more details.

\paragraph{Optimization}
We sample random batches of training rays and optimize our color and distance fields by minimizing the photometric loss $\mathcal{L}_{\text{photo}}$ and Eikonal loss $\mathcal{L}_{\text{Eik}}$ along with interlevel loss $\mathcal{L}_{\text{prop}}$ \cite{BarroMVSH2022} to train the proposal network:
\begin{equation}
    \mathcal{L}(\vct r) \coloneqq \mathcal{L}_{\text{photo}}(\vct r) + \lambda_{\text{Eik}} \mathcal{L}_{\text{Eik}}(\vct r) + \mathcal{L}_{\text{prop}}(\vct r) \text{,}
    \label{eq:training-loss}
\end{equation}
with $\lambda_{\text{Eik}} = 0.01$ in our experiments.

\subsection{Finetuning}
\label{sec:finetuning}

\paragraph{Adaptive surfaceness}

The first stage of our pipeline uses a global surfaceness value $\beta(\vct{x}) = \bar\beta$ for all $\vct x$, as in existing approaches \cite{WangLLTKW2021,YarivGKL2021}.
As $\bar\beta$ increases, the density $\sigma_\text{SDF}$ in free-space areas converges to zero (\cref{eq:sdf-density}), reducing the required number of samples per ray.
However, uniformly increasing this scene-wide parameter degrades the rendering quality near fine-grained and transparent structures (see \cref{fig:different-beta}).

\begin{figure}
    \captionsetup[subfigure]{justification=centering}
        \centering
        \begin{subfigure}[b]{0.49\linewidth}
            \centering
		\caption*{\textbf{RGB}}
            \includegraphics[width=\textwidth]{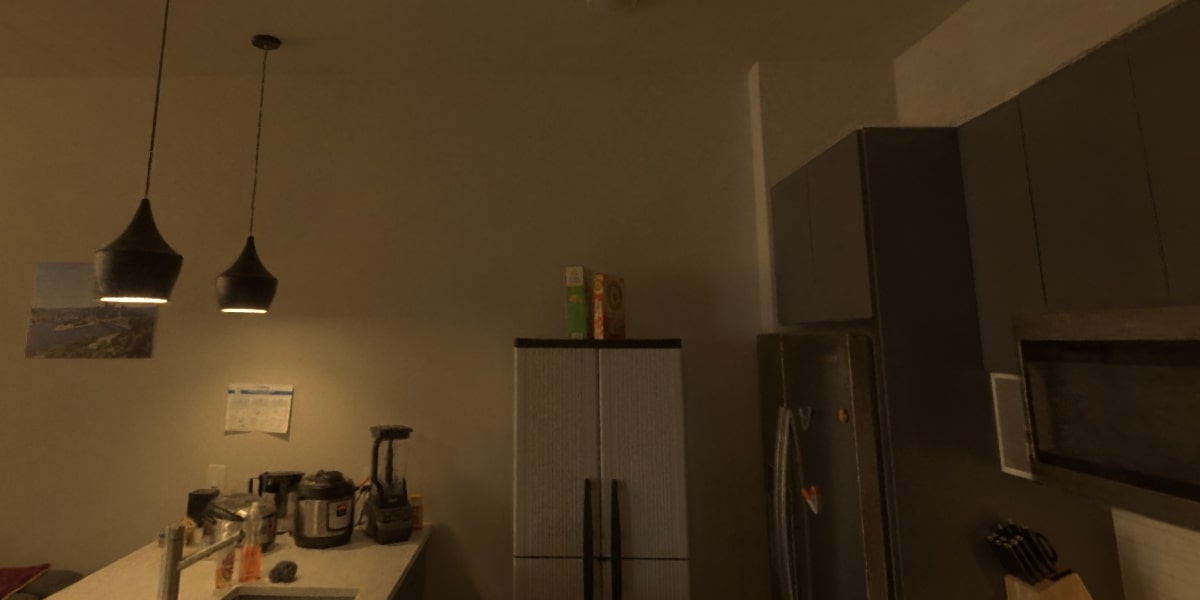}
        \end{subfigure}
        \hfill
        \begin{subfigure}[b]{0.49\linewidth}  
            \centering 
		\caption*{\textbf{Eikonal Loss}}
            \includegraphics[width=\textwidth]{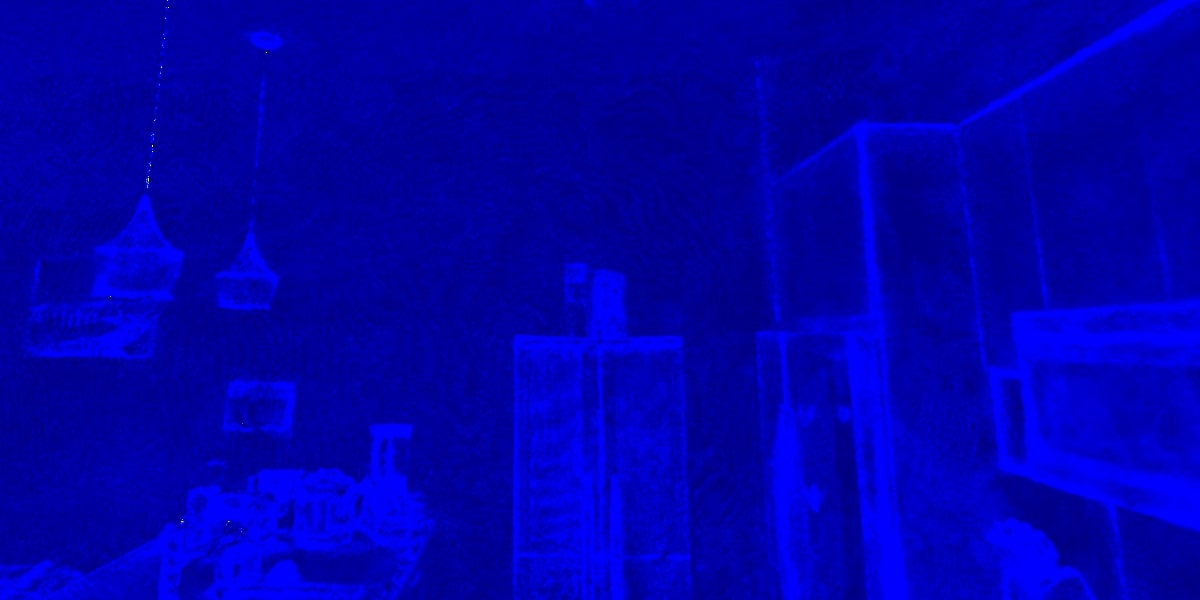}
        \end{subfigure}
        \smallskip
        \begin{subfigure}[b]{0.49\linewidth}   
            \centering 
		\caption*{\textbf{Surfaceness}}
            \includegraphics[width=\textwidth]{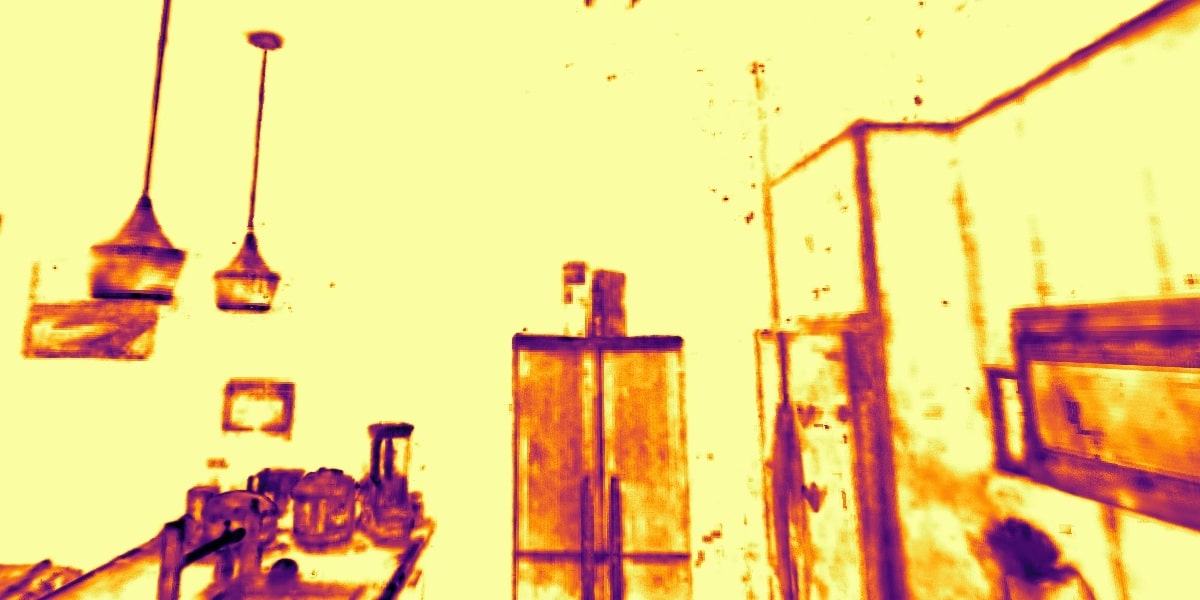}
        \end{subfigure}
        \caption{\textbf{Spatially adaptive surfaceness.} 
        We make $\beta(\vct{x})$ spatially adaptive by means of a $512^3$ voxel grid that we increase during the finetuning stage.
		We track Eikonal loss as we increase surfaceness as it is highest near object boundaries and semi-transparent surfaces (\textbf{top-right}, brighter = higher loss) that degrade when surfaceness is too high (\cref{fig:different-beta}).
		We stop increasing surfaceness in regions that cross a given threshold.
	}
\label{fig:localized-betas}
\vspace*{-4mm}
\end{figure}

We overcome this limitation by making $\beta(\vct{x})$ spatially adaptive via a $512^3$ voxel grid.
One possible approach is to directly optimize $\beta(\vct{x})$ via gradient descent, but we find that this overly relaxes the constraint on SDF correctness such that $f(\vct x)$ predicts arbitrary density values as in the original NeRF.
We instead rely on the Eikonal loss as a natural indicator of where the model cannot accurately reconstruct the scene via an SDF (and where we should therefore use a ``softer” formulation).
We collect per-sample triplets $(\vct x, \eta, w)$ rendered during the finetuning process, accumulate them over multiple training iterations (5,000), 
and partition them across the voxels of the surfaceness grid.
Let $\Lambda_{\vct v}$ be the subset associated with voxel $\vct v$ corresponding to $\beta_{\vct v}$.
We increase $\beta_{\vct v}$ by a fixed increment (100) if:

\begin{equation}
\label{eq:surfaceness-formula}
\frac{\sum_{(\vct x, \eta, w) \in \Lambda_{\vct v}} w \eta(\Vert\nabla f(\vct x)\Vert - 1)^2}{\sum_{(\ldots, w) \in \Lambda_{\vct v}} w} < \bar\gamma \text{,}
\end{equation}
where $\bar\gamma \coloneqq 0.25$ is a predefined threshold.
\cref{fig:localized-betas} illustrates our approach.

\paragraph{Proposal network baking}

Although the proposal network allows us to quickly learn the scene geometry during the first stage of training, it is too expensive to evaluate in real time.
We follow MERF's protocol \cite{ReiseSVSMGBH2023} to bake the proposal network into a $1024^3$ binary occupancy grid.
We render all training rays and mark a voxel as occupied if there exists at least one sampled point $\vct x_i$ such that $\max(w_i,\sigma_i)>0.005$.
We finetune our model using the occupancy grid to prevent any loss in quality.

\paragraph{MLP distillation}
We find it important to use a large 256 channel-wide MLP to represent the signed distance $f$ during the first training phase in order to learn accurate scene geometry.
However, we later distill $f$ into a smaller 16-wide network ($f_\text{small})$.
We do so by sampling random rays from our training set for 5,000 iterations and minimizing the difference between $f(\vct x_i)$ and $f_\text{small}(\vct x_i)$ at every sampled point:
\begin{equation}
    \mathcal{L}_{\text{dist}}(\vct r) \coloneqq \sum_{i=0}^{N-1} |f(\vct{x}_i) - f_\text{small}(\vct{x}_i)|,
    \label{eq:distill-loss}
\end{equation}
with a stop gradient applied to the outputs of $f$.
We then discard the original SDF $f$ and switch to using the distilled counterpart $f_\text{small}$ for the rest of the finetuning stage.

\subsection{Backgrounds}
\label{sec:backgrounds}

Many scenes we wish to reconstruct contain complex backgrounds that surface--volume methods struggle to replicate \cite{YarivGKL2021, WangLLTKW2021, LiMETULL2023}.
BakedSDF \cite{YarivHRVSSBM2023} defines a contraction space \cite{BarroMVSH2022} in which the Eikonal loss of \cref{eq:eikonal-loss1} is applied.
However, we found this to negatively impact foreground quality.
Other approaches use separate NeRF background models \cite{ZhangRSK2020}, which effectively doubles inference and memory costs, and makes them ill-suited for real-time rendering.

\paragraph{Relation between volumetric and surface-based NeRFs}

We discuss how to make a single MLP behave as an approximate SDF in the foreground and a volumetric model in the background.
Both types of NeRF derive density $\sigma$ by applying a non-linearity to the output of an MLP.
Our insight is that although the original NeRF uses ReLU, any non-linear mapping to $\mathbb{R}^+$ may be used in practice, including our scaled CDF $\Psi$ ($\beta$ omitted without loss of generality).
Since $\Psi$ is invertible (as it is a CDF), $\sigma(\vct{x})$ and $\Psi(f(\vct{x}))$ are functionally equivalent as there exists an $f$ such that $\Psi(f(\vct{x}))$ = $\sigma(\vct{x})$ for any given point $\vct{x}$.
Put otherwise, it is the Eikonal regularization that causes the divergence in behavior between both methods --- in its absence, an ``SDF" MLP is free to behave exactly
as the density MLP in the original NeRF!

\begin{figure}
    \captionsetup[subfigure]{justification=centering}
        \centering
        \begin{subfigure}[b]{0.49\linewidth}
            \centering
		\caption*{\textbf{Eikonal Loss ($\eta_i = 1$)}}
            \includegraphics[width=\textwidth]{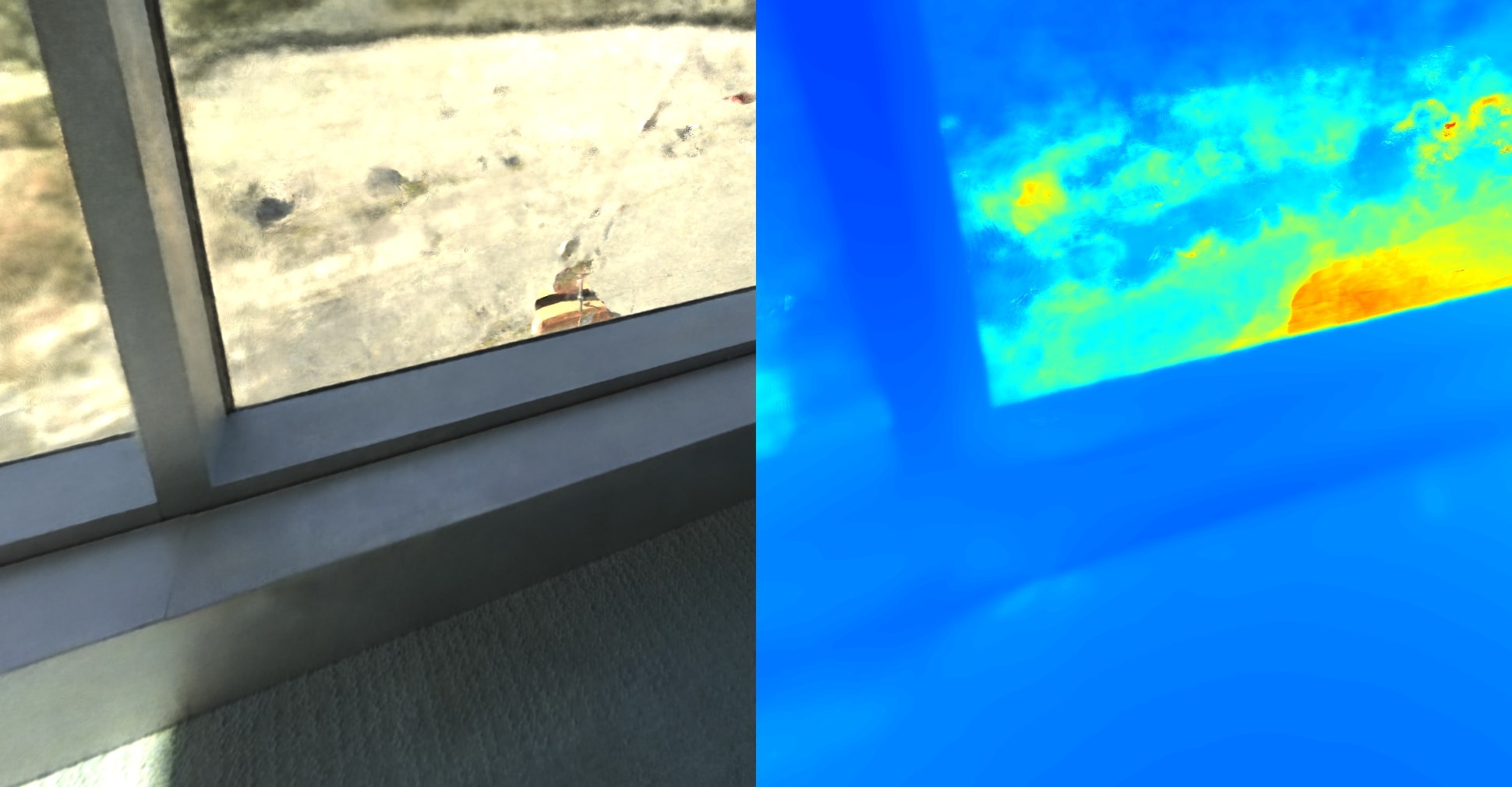}
        \end{subfigure}
        \hfill
        \begin{subfigure}[b]{0.49\linewidth}  
            \centering 
		\caption*{\textbf{No Eikonal Loss}}
            \includegraphics[width=\textwidth]{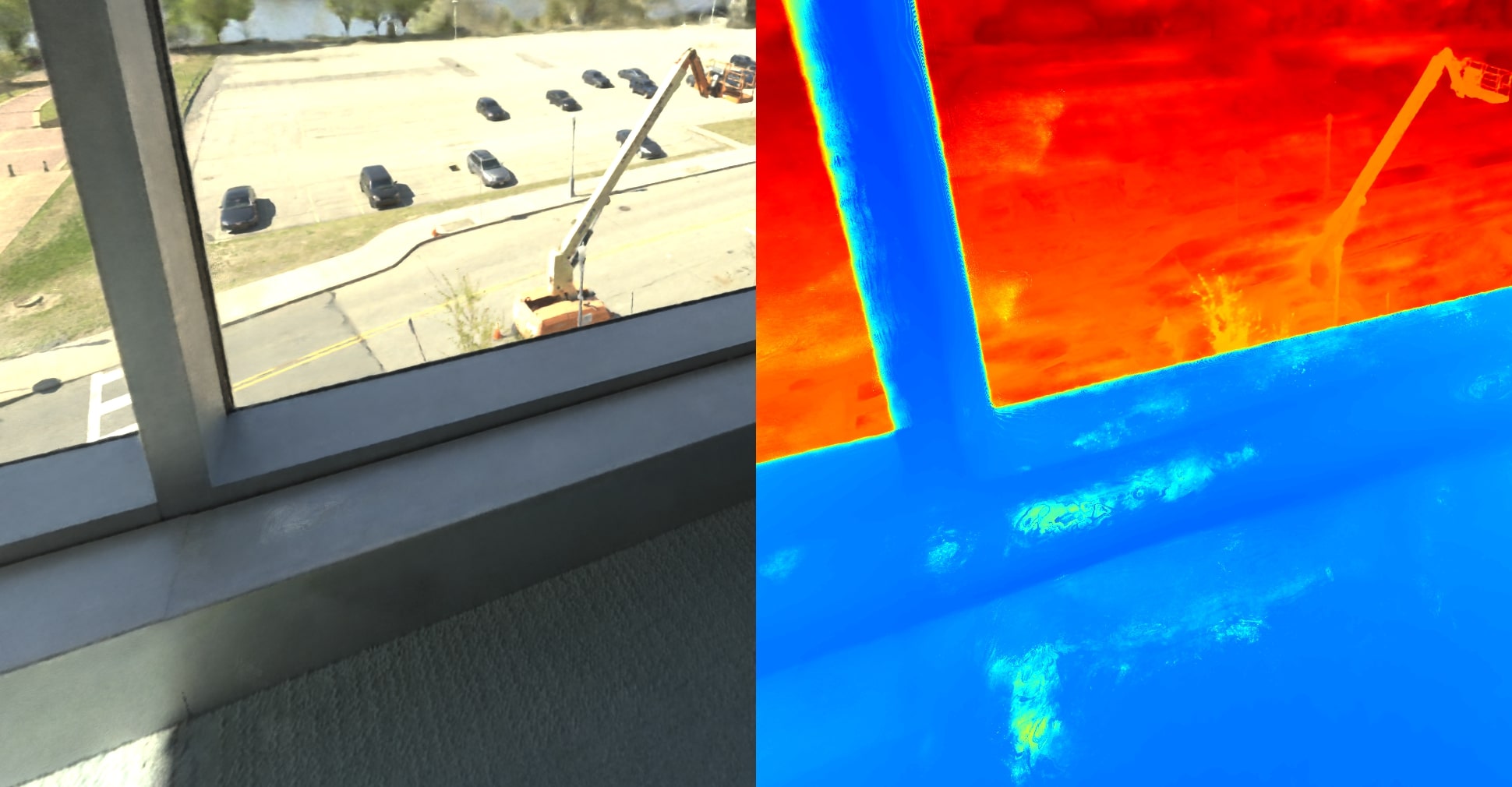}
        \end{subfigure}
        \smallskip
        \begin{subfigure}[b]{0.49\linewidth}   
            \centering 
		\caption*{\textbf{Eikonal Loss ($\eta_i = 1$) in Contracted Space \cite{YarivHRVSSBM2023}}}
            \includegraphics[width=\textwidth]{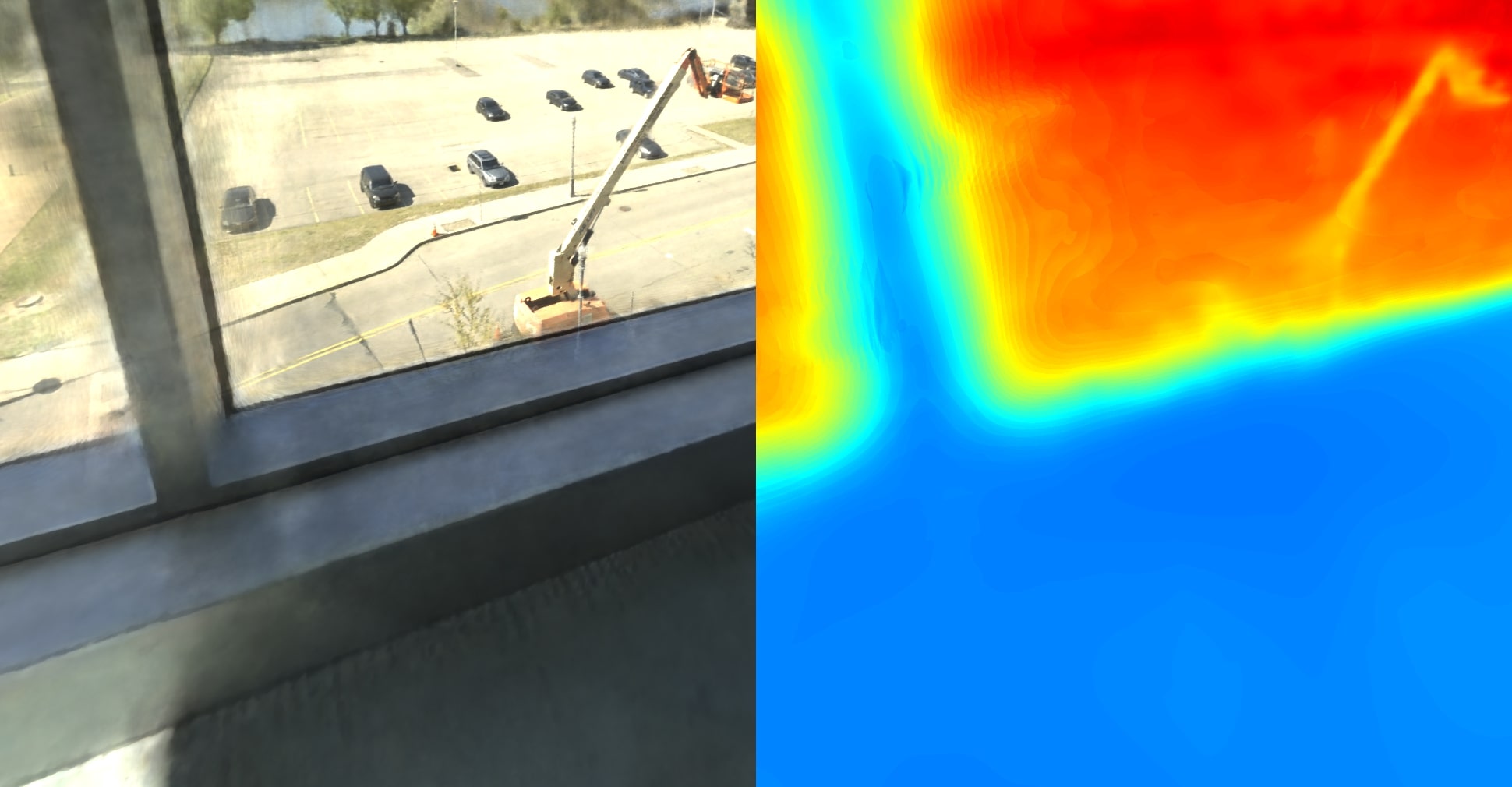}
        \end{subfigure}
        \hfill
        \begin{subfigure}[b]{0.49\linewidth}   
            \centering 
            \caption*{\textbf{Distance-Adjusted Eikonal Loss ($\eta_i = d_i^{-2}$)}}
            \includegraphics[width=\textwidth]{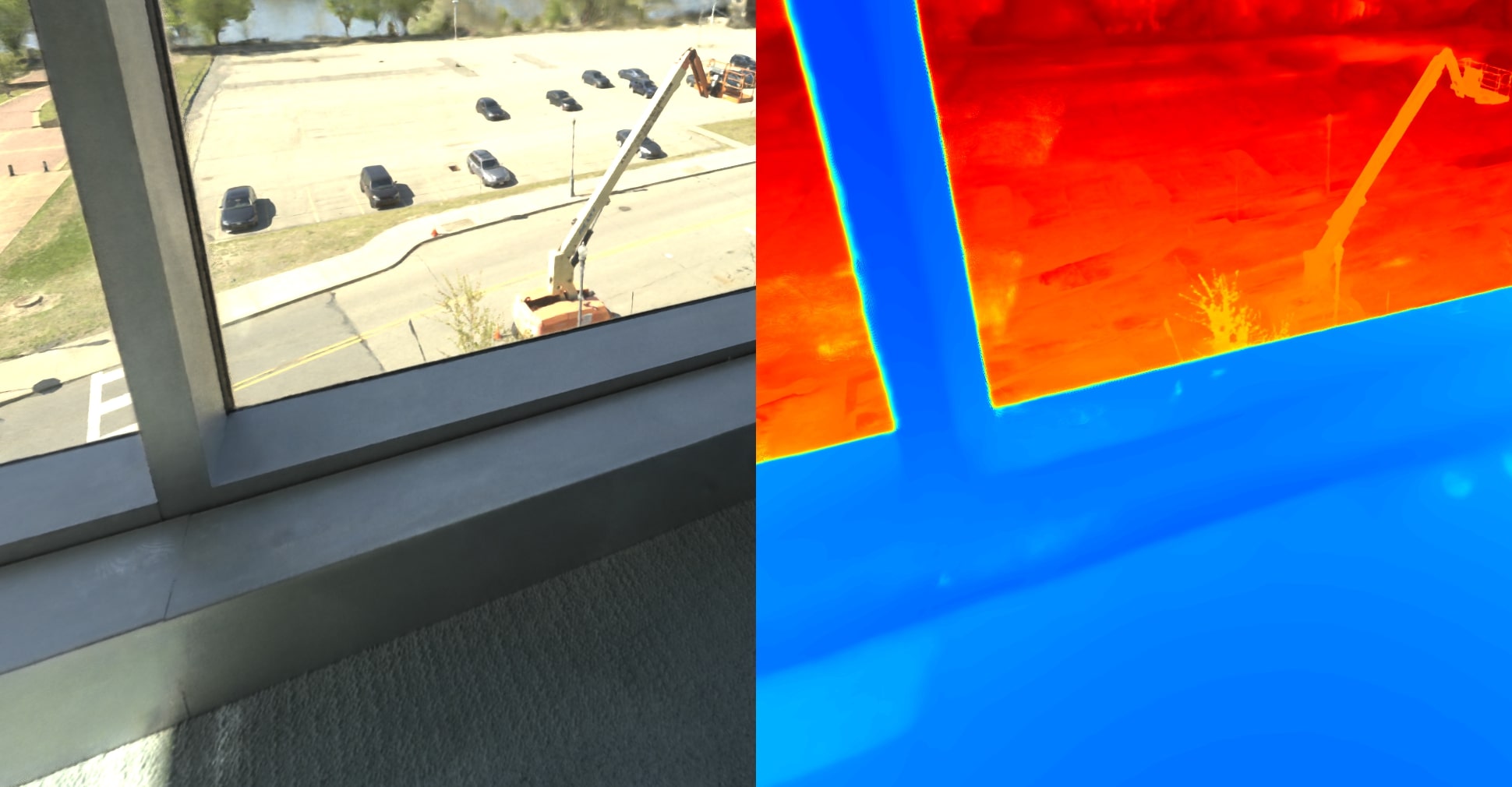}
        \end{subfigure}
	\caption{\textbf{Backgrounds.}
            Using standard Eikonal loss affects background reconstruction (\textbf{top-left}) while applying it in contracted space \cite{YarivHRVSSBM2023} affects the foreground (\textbf{bottom-left}).
            Omitting Eikonal loss entirely causes surface--volume methods to revert to NeRF's behavior, which improves background quality but degrades foreground surface reconstruction (\textbf{top-right}).
            By using distance-adjusted sample weights $\eta_i = d_i^{-2}$, we improve background reconstruction without impacting foreground quality (\textbf{bottom-right}).
	}
	\label{fig:eikonal-scaling}
\vspace*{-4mm}
\end{figure}

\paragraph{Distance-adjusted loss}

We use a \textit{distance-adjusted} Eikonal loss during training by using per-sample loss weights $\eta_i = \frac{1}{d_i^2}$ (where $d_i$ is the metric distance along the ray of sample $\vct x_i$) instead of commonly-used uniform weights ($\eta_i = 1$) to downweight the loss applied to far-field regions.
Intuitively, this encourages our method to behave as a valid SDF in the foreground (with well-defined surfaces) and more like NeRF in the background (to enable accurate reconstruction)
without the need for separate foreground and background models.
\cref{fig:eikonal-scaling} and \cref{table:depth-results} illustrate the different approaches.

\subsection{Real-Time Rendering}
\label{sec:rendering}

\paragraph{Texture storage}

Our architecture enables us to use lower-level optimizations.
Methods such as iNGP \cite{MuelleESK2022} use concatenated multi-resolution features stored in hash tables.
Since we use explicit 3D grids and triplanes, we can store our features as textures at render time,
taking advantage of increased memory locality and texture interpolation hardware.
As we sum our multi-resolution features during training, we optimize the number of texture fetches by storing pre-summed features $g'$ at resolution level $L$ (where we store $g'(\vct{v}) = \sum_{l=0}^{L} g(\vct{v}, l)$ for each texel in $L$).
For a given sample $\vct{x}$ at render time, we obtain its anti-aliased feature by interpolating between the two levels implied by its pixel area $p(\vct{x})$, reducing the number of texture fetches to 8 queries per MLP evaluation from the original $3 \!+\! 3 \!\times\! 7 \!=\! 24$ (assuming three 3D grids and seven triplane levels), a 3$\times$ reduction.

\begin{figure*}[t!]
    \centering
    \includegraphics[width=\textwidth, clip, trim = 0mm 2mm 0mm 0mm]{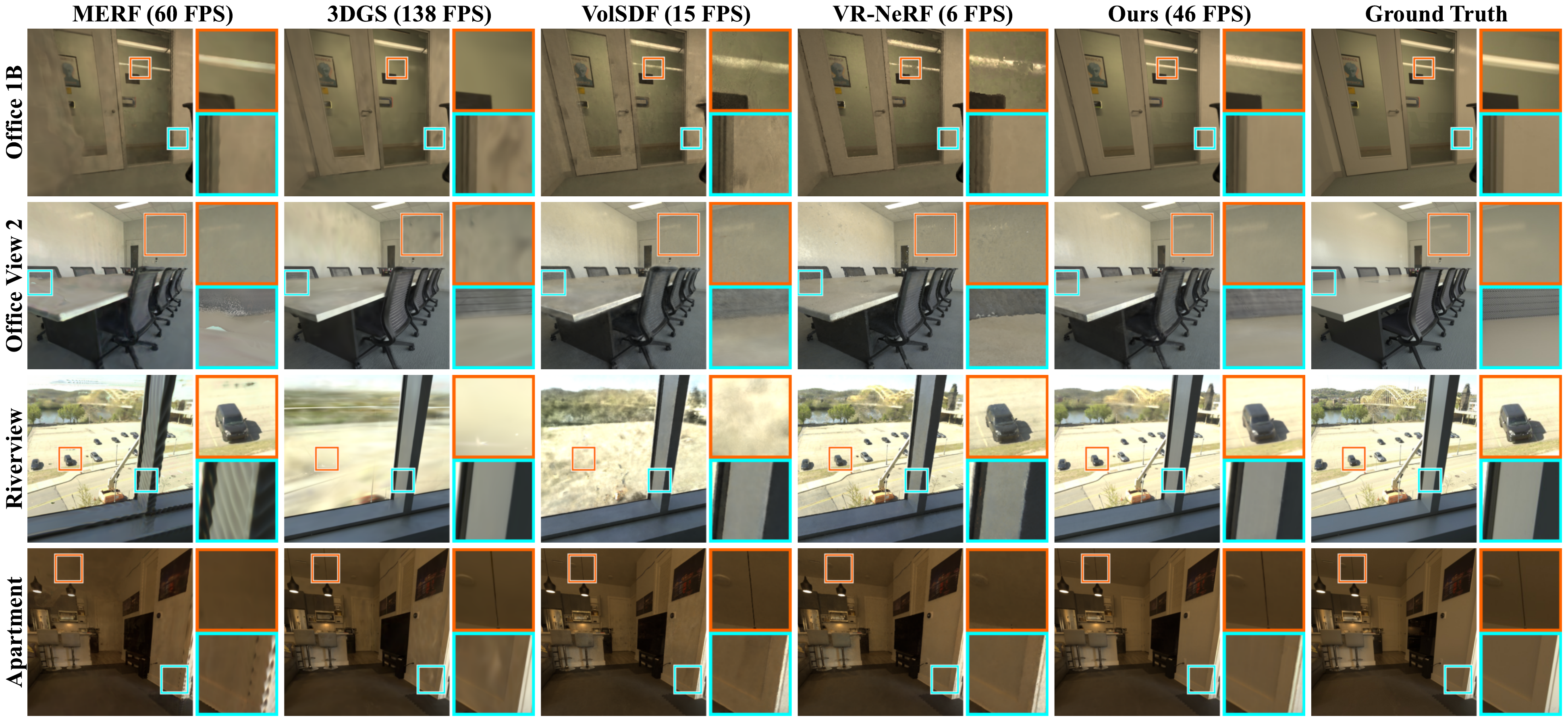}
    \caption{\textbf{Eyeful Tower \cite{XuALGBKRPKBLZR2023}.}
        \method is the only method to accurately model reflections and shadows (\textbf{first two rows}), far-field content (\textbf{third row}) and fine structures (\textbf{bottom row}) at real-time frame rates at 2K$\times$2K resolution.
    }
    \label{fig:eyeful-qualitative}
    \vspace*{-2mm}
\end{figure*}

\begin{table*}
    \caption{\textbf{Eyeful Tower \cite{XuALGBKRPKBLZR2023} results.}
        We omit 3DGS results for fisheye scenes as their implementation does not handle fisheye projection.
        Along with 3DGS and MERF, ours is the only to reach the 36 FPS target for VR along with a $>$1.5\,dB PSNR improvement in quality.
    }
	\centering
	\resizebox{\linewidth}{!}{%
		\begin{tabular}{l@{\hspace{1em}}c@{\hspace{1em}}c@{\hspace{1em}}c@{\hspace{2em}}c@{\hspace{1em}}c@{\hspace{1em}}c@{\hspace{2em}}c@{\hspace{1em}}c@{\hspace{1em}}c@{\hspace{1em}}r}
		\toprule
		& \multicolumn{3}{c@{\hspace{2em}}}{Pinhole}
		& \multicolumn{3}{c@{\hspace{2em}}}{Fisheye}
		& \multicolumn{4}{c}{Overall}
		\\ \cmidrule(r{2em}){2-4}\cmidrule(r{2em}){5-7}\cmidrule(r{0.6em}){8-11}
		& $\uparrow$PSNR & $\uparrow$SSIM & $\downarrow$LPIPS
		& $\uparrow$PSNR & $\uparrow$SSIM & $\downarrow$LPIPS
		& $\uparrow$PSNR & $\uparrow$SSIM & $\downarrow$LPIPS & $\uparrow$FPS \\ \midrule
		iNGP* \cite{MuelleESK2022} & 27.35 & 0.826 & 0.361 & 33.32 & 0.938 & 0.155 & 30.06 & 0.877 & 0.267 & 4.55 \\
		VolSDF* \cite{YarivGKL2021} & 27.10 & 0.856 & 0.310 & 34.09 & 0.951 & 0.116 & 30.28 & 0.899 & 0.222 & 15.29 \\
		MERF (pre-baking) \cite{ReiseSVSMGBH2023} & 26.44 & 0.831 & 0.506 & 31.18 & 0.922 & 0.549 & 28.59 & 0.872 & 0.526 & 18.11 \\
		MERF (baked)~\cite{ReiseSVSMGBH2023} & 25.99 & 0.830 & 0.525 & 31.09 & 0.921 & 0.546 & 28.31 & 0.871 & 0.535 & \underline{60.18} \\
		3D Gaussian splatting~\cite{KerblKLD2023}     & 27.42 & \underline{0.877} & \underline{0.291} 
  &  ---  &  ---  &  --- &  ---  &  ---  &  ---  & \textbf{138.22} \\
		VR-NeRF~\cite{XuALGBKRPKBLZR2023} & \underline{28.08} & 0.834 & 0.326 & \underline{34.53} & 0.951 & 0.130 & 31.01 & 0.888 & 0.237 & 6.05 \\
		Zip-NeRF~\cite{BarroMVSH2023} & \textbf{29.71} & 0.868 & 0.305 & 34.19 & \textbf{0.958} & \textbf{0.109} & \textbf{31.75} & \underline{0.909} & \underline{0.216} & \textless 0.1 \\
  \midrule
		\method & \underline{29.07} & \textbf{0.880} & \textbf{0.268} & \textbf{34.57} & \underline{0.952} & \underline{0.115} & \underline{31.57} & \textbf{0.913} & \textbf{0.198} & 45.78 \\
		\end{tabular}%
	}
    \\[3pt]{\footnotesize * Our implementation. VolSDF: with iNGP acceleration.}
\vspace*{-4mm}
	\label{table:eyeful-results}
\end{table*}

\paragraph{Sphere tracing}

Volumetric methods that use occupancy grids \cite[e.g.][]{MuelleESK2022, ReiseSVSMGBH2023} sample within occupied voxels using a given step size.
This hyperparameter must be carefully tuned to strike the proper balance between quality (not skipping thin surfaces) and performance (not excessively sampling empty space).
Modeling an SDF allows us to sample more efficiently by advancing toward the predicted surface using \emph{sphere tracing} \cite{RosuB2023}.
At each sample point $\vct{x}_i$ and predicted surface distance $s \!=\! f(\vct{x}_i)$, we advance by $0.9s$ (chosen empirically to account for our model behaving as an approximate SDF) until hitting the surface (predicted as $s \!\leq\! 2 \!\times\! 10^{-4}$).
We only perform sphere tracing where our model behaves as a valid SDF
(determined by $\beta(\vct{x}_i) \!>\! 350$ in our experiments),
and fall back to a predefined step size of 1\,cm otherwise.

\section{Experiments}
\label{sec:experiments}

As our goal is high-fidelity view synthesis at VR resolution ($\approx$ 4 megapixels), we primarily evaluate \method against the Eyeful Tower dataset \cite{XuALGBKRPKBLZR2023}, which contains high-fidelity scenes designed for walkable VR (\cref{sec:eyeful-eval}).
We compare our work to a broader range of methods on additional datasets in \cref{sec:common-eval}.
We ablate our design in \cref{sec:diagnostics}.

\subsection{Implementation}

We train our models in the PyTorch framework \cite{PaszkGMLBCKLGADKYDRTCSFBC2019} and implement our renderer in C++/CUDA.
We parameterize unbounded scenes with MERF's piecewise-linear contraction \cite{ReiseSVSMGBH2023} so that our renderer can query the occupancy grid via ray-AABB intersection.
We train on each scene for 200,000 iterations (100,000 in each training stage) with 12,800 rays per batch using Adam \cite{KingmB2015} and a learning rate of $2.5 \!\times\! 10^{-3}$.

\begin{figure*}[t!]
    \centering
    \includegraphics[width=\textwidth, clip, trim = 0mm 2mm 0mm 0mm]{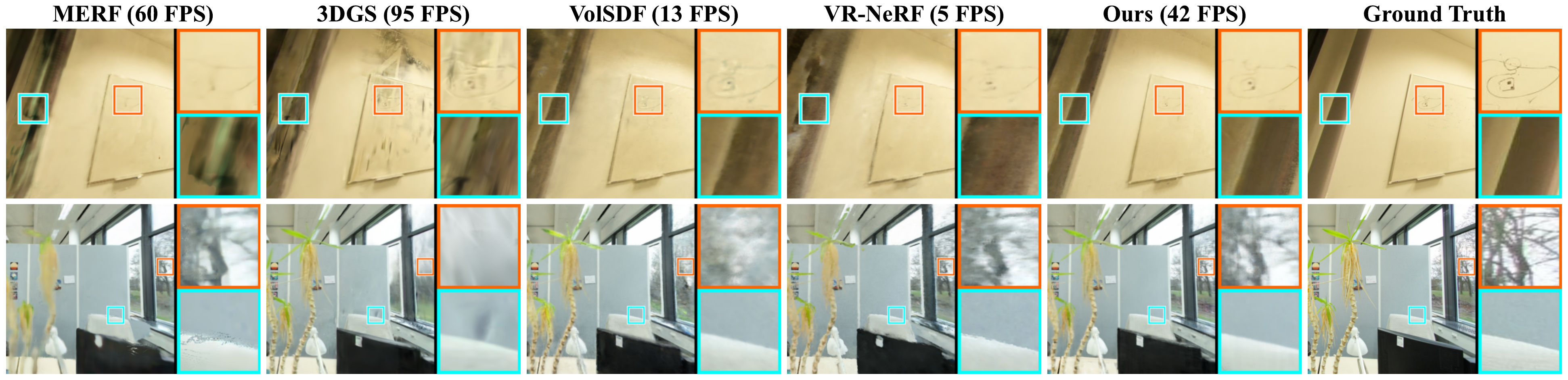}
    \caption{\textbf{ScanNet++ \cite{YeshwLND2023}.}
        3D Gaussian splatting \cite{KerblKLD2023} struggles with specular surfaces such as whiteboards (\textbf{above}) and far-field content (\textbf{below}).
        Our method performs best qualitatively while maintaining a real-time frame rate.}
    \label{fig:scannet-qualitative}
    \vspace*{-2mm}
\end{figure*}

\sethlcolor{realtimecolor}

\begin{table*}
    \caption{\textbf{MipNeRF 360 \cite{BarroMVSH2022}.}
        Real-time methods are \hl{highlighted} (\textbf{best}, \underline{second-best}, \textit{third-best}).
        Baseline numbers as published \cite{ReiseSVSMGBH2023, YarivHRVSSBM2023, ChenFHT2023}.
        MobileNeRF \cite{ChenFHT2023} was not evaluated on indoor scenes.
        Our method performs similar to state-of-the-art real-time and offline methods.
	}
	\centering
	\resizebox{\linewidth}{!}{%
		\begin{tabular}{l@{\hspace{1em}}c@{\hspace{1em}}c@{\hspace{1em}}c@{\hspace{2em}}c@{\hspace{1em}}c@{\hspace{1em}}c@{\hspace{2em}}c@{\hspace{1em}}c@{\hspace{1em}}c}
		\toprule
		& \multicolumn{3}{c@{\hspace{2em}}}{Outdoor}
		& \multicolumn{3}{c@{\hspace{2em}}}{Indoor}
		& \multicolumn{3}{c}{Overall}
		\\ \cmidrule(r{2em}){2-4}\cmidrule(r{2em}){5-7}\cmidrule(r{0.6em}){8-10}
		& $\uparrow$PSNR & $\uparrow$SSIM & $\downarrow$LPIPS
		& $\uparrow$PSNR & $\uparrow$SSIM & $\downarrow$LPIPS
		& $\uparrow$PSNR & $\uparrow$SSIM & $\downarrow$LPIPS \\ \midrule
		NeRF~\cite{mildenhall2020nerf} & 21.46 & 0.458 & 0.515 & 26.84 & 0.790 & 0.370 & 23.85 & 0.606 & 0.451  \\
		NeRF++~\cite{ZhangRSK2020} &  22.76 & 0.548 & 0.427 & 28.05 & 0.836 & 0.309 & 25.11	& 0.676 & 0.375 \\
		SVS~\cite{Riegler2021SVS} & 23.01 & 0.662 & 0.253 & 28.22 & 0.907 & 0.160 & 25.33 & 0.771 & 0.212 \\
		Mip-NeRF 360~\cite{BarroMVSH2022} & \textit{24.47} & 0.691 & 0.283 & \underline{31.72} & 0.917 & 0.180 & \underline{27.69} & 0.791 & 0.237 \\
		iNGP~\cite{MuelleESK2022} & 22.90 & 0.566 & 0.371 & 29.15 & 0.880 & 0.216 & 25.68 & 0.706 & 0.302 \\
            Zip-NeRF~\cite{BarroMVSH2023} & \textbf{25.46} & \textbf{0.747} & \textbf{0.170} & \textbf{32.29} & \textbf{0.931} & \textit{0.106} & \textbf{28.49} & \textbf{0.829} & \textbf{0.142} \\
		\midrule
		\rowcolor{realtimecolor} Deep Blending~\cite{hedman_deepblending} & 21.54 & 0.524 & 0.364 & 26.40 & 0.844 & 0.261 & 23.70 & 0.666 & 0.318 \\ 
		\rowcolor{realtimecolor} MobileNeRF~\cite{ChenFHT2023} & 21.95 & 0.470 & 0.470 & --- & --- & --- & --- & --- & --- \\
		\rowcolor{realtimecolor} BakedSDF~\cite{YarivHRVSSBM2023} & 22.47 & 0.585 & 0.349 & 29.15 & 0.880 & 0.216 & 25.68 & 0.706 & 0.302 \\
		\rowcolor{realtimecolor} MERF~\cite{ReiseSVSMGBH2023} & 23.19 & 0.616 & 0.343 & 27.80 & 0.855 & 0.271 & 25.24 & 0.722 & 0.311 \\
		\rowcolor{realtimecolor} 3D Gaussian splatting~\cite{KerblKLD2023} & 24.13 & \textit{0.707} & \underline{0.211} & 30.94 & \underline{0.927} & \textbf{0.081} & 27.16 & \textit{0.805} & \underline{0.153} \\
		\midrule
		\rowcolor{realtimecolor} \method & \underline{24.73} & \underline{0.716} & \textit{0.224} &	\textit{31.01} & \textit{0.920} & \underline{0.095} & \textit{27.52} & \underline{0.806} & \textit{0.167} \\
		\bottomrule
		\end{tabular}%
	}
	\label{table:m360-results}
 \vspace*{-4mm}
\end{table*}

\subsection{VR Rendering}
\label{sec:eyeful-eval}

\paragraph{Eyeful Tower dataset}

The dataset consists of room-scale captures, each containing high-resolution HDR images at 2K resolution, captured using a multi-view camera rig.
Although care is taken to obtain the best quality images possible, inconsistencies still appear between images due to lighting changes and shadows from humans and the capture rig itself.
We model as much of the dynamic range as possible by mapping colors in the PQ color space \cite{SMPTE2084-2014},
as proposed in VR-NeRF \cite{XuALGBKRPKBLZR2023}, during training and tonemap to sRGB space during evaluation to compare against non-HDR baselines.

\paragraph{Baselines}

We compare \method\ to baselines across the fidelity/speed spectrum.
We benchmark several volumetric methods, including
(1) iNGP \cite{MuelleESK2022},
(2) VR-NeRF \cite{XuALGBKRPKBLZR2023}, which extends iNGP's \cite{MuelleESK2022} primitives to better handle HDR reconstruction, 
(3) Zip-NeRF \cite{BarroMVSH2023}, an anti-aliasing method that generates high-quality renderings at the cost of speed, and
(4) MERF \cite{ReiseSVSMGBH2023}, a highly optimized method that uses sampling and memory layout optimizations to accelerate rendering.
We also compare to VolSDF \cite{YarivGKL2021} as a hybrid surface--volume method similar to the first stage of our method.
As the original VolSDF implementation uses large MLPs that are unsuitable for real-time rendering, we use an optimized version built on top of iNGP's acceleration primitives as a fairer comparison.
We also benchmark 3D Gaussian splatting \cite{KerblKLD2023} as a non-neural approach that represents the current state of the art with across rendering quality and speed.

\paragraph{Metrics}

We report quantitative results based on PSNR, SSIM \cite{ssim}, and the AlexNet implementation of LPIPS \cite{zhang2018perceptual} and measure frame rates rendered at 2K$\times$2K resolution on a single NVIDIA RTX 4090 GPU.

\paragraph{Results}

We summarize our results in \cref{table:eyeful-results} along with qualitative results in \cref{fig:eyeful-qualitative}.
VR-NeRF \cite{XuALGBKRPKBLZR2023}, iNGP \cite{MuelleESK2022}, and Zip-NeRF \cite{BarroMVSH2023} render well below real-time frame rates.
Our VolSDF implementation, which uses the same primitives as iNGP, is 3$\times$ faster merely from the benefits of using a surface representation (and fewer samples per ray).
MERF \cite{ReiseSVSMGBH2023}, as a volume representation, relies instead on precomputation to accelerate rendering by explicitly storing diffuse color and density outputs during its baking stage and using only a small MLP to model view-dependent effects.
Although it reaches a high frame rate, it provides the least visually appealing results amongst our baselines.
3D Gaussian splatting \cite{KerblKLD2023} renders the fastest, but struggles with shadows and lighting changes across the training views and models them as unsightly floaters.
Our method is the only to achieve both high quality and real-time frame rates.

\begin{table*}
    \caption{\textbf{Diagnostics.}
        A global learned $\beta$ ($\approx 200$) produces the highest-quality renderings, but is slow to render as much of the scene is modeled volumetrically.
        Increasing $\beta$ improves rendering speed but results in worse accuracy.
        Our full method (with spatially-varying $\beta(\vct{x})$) gets the best of both worlds.
        Other innovations such as distance-adjusted Eikonal loss are crucial for ensuring high accuracy for scenes with complex backgrounds.
        Finally, distillation and hardware acceleration come at a minor quality cost while doubling rendering speed.
    }
	\centering
		\begin{tabular}{l||cccc|cccc}
		\toprule
		Methods & 
		$\beta(\vct{x})$ &
            Dist. &
		Distill &
		Textures &
		$\uparrow$PSNR &
		$\uparrow$SSIM & 
		$\downarrow$LPIPS &
		$\uparrow$FPS  \\ \midrule

		w/ Global $\beta$ (learned) & \xmark & \textcolor{CheckGreen}{\checkmark} & \textcolor{CheckGreen}{\checkmark} & \textcolor{CheckGreen}{\checkmark} 
		& \textbf{31.76} & \textbf{0.923} & \textbf{0.188} & 28.79  \\
  	w/ Global $\beta$ = 2000 & \xmark  & \textcolor{CheckGreen}{\checkmark} & \textcolor{CheckGreen}{\checkmark} & \textcolor{CheckGreen}{\checkmark} 
		& 27.16 & 0.835 & 0.345	& \textbf{47.47}  \\
		w/o distance-adjusted Eik. & \textcolor{CheckGreen}{\checkmark} & \xmark & \textcolor{CheckGreen}{\checkmark} & \textcolor{CheckGreen}{\checkmark} 
		& 29.97 & 0.856 & 0.260 & 45.42 \\
  w/o MLP Distillation & \textcolor{CheckGreen}{\checkmark} & \textcolor{CheckGreen}{\checkmark} & \xmark & \textcolor{CheckGreen}{\checkmark} 
		& \underline{31.65} & 0.915 & \underline{0.193} & 35.25 \\
		w/o CUDA Textures & \textcolor{CheckGreen}{\checkmark} & \textcolor{CheckGreen}{\checkmark} & \textcolor{CheckGreen}{\checkmark} & \xmark
		& 31.62 & \underline{0.921} & 0.195 & 28.48 \\
		\midrule
		Full Method & \textcolor{CheckGreen}{\checkmark} & \textcolor{CheckGreen}{\checkmark} & \textcolor{CheckGreen}{\checkmark} & \textcolor{CheckGreen}{\checkmark} 
		& 31.57 & 0.913 & 0.198 & \underline{45.78}  \\
		\end{tabular}
\label{table:diagnostics}
\vspace*{-4mm}
\end{table*}

\subsection{Additional Comparisons}
\label{sec:common-eval}

\paragraph{Datasets}

We evaluate \method on MipNeRF-360 \cite{BarroMVSH2022} as a highly-referenced dataset evaluated by many prior methods, and ScanNet++ \cite{YeshwLND2023} as a newer benchmark built from high-resolution captures of indoor scenes that are relevant to our goal of enabling immersive AR/VR applications.
We test on all scenes in the former and a subset of the latter.

\paragraph{Baselines}

We compare \method to a wide set of baselines on Mip-NeRF 360.
We use the same set of baselines as in \cref{sec:eyeful-eval} for ScanNet++.

\paragraph{Results}

We list results in \cref{table:m360-results} and \cref{table:snpp-results}.
Our method performs comparably to the best on Mip-NeRF 360 across both real-time \cite{KerblKLD2023} and offline \cite{BarroMVSH2022} methods.
Although ScanNet++ \cite{YeshwLND2023} contains fewer lighting inconsistencies across training images than the Eyeful Tower dataset \cite{XuALGBKRPKBLZR2023}, 3D Gaussian splatting still struggles to reconstruct specular surfaces (whiteboards, reflective walls) and backgrounds (\cref{table:snpp-results}).
Our method performs the best amongst real-time methods and comparably to Zip-NeRF \cite{BarroMVSH2023}, while rendering \textgreater 400$\times$ faster.

\begin{table}
    \caption{\textbf{ScanNet++ \cite{YeshwLND2023} results.}
    Similar to \cref{table:eyeful-results}, our method is the only to hit VR FPS rates along with 3DGS and MERF.
    Our quality is near-identical to Zip-NeRF while rendering \textgreater 400$\times$ faster.}
	\centering
	\resizebox{\linewidth}{!}{%
		\begin{tabular}{l@{\hspace{1em}}c@{\hspace{1em}}c@{\hspace{1em}}c@{\hspace{1em}}r}
		\toprule 
		Method & $\uparrow$PSNR & $\uparrow$SSIM & $\downarrow$LPIPS & $\uparrow$FPS \\ \midrule
		iNGP*~\cite{MuelleESK2022} & 23.69 & 0.815 & 0.308 & 5.39 \\
		VolSDF*~\cite{YarivGKL2021} & 24.26 & 0.834 & 0.246 & 13.18 \\
		MERF (pre-baking)~\cite{ReiseSVSMGBH2023} & 23.44 & 0.821 & 0.306 & 12.08 \\
		MERF (baked)~\cite{ReiseSVSMGBH2023} & 23.19 & 0.820 & 0.308 & \underline{60.21} \\
            3D Gaussian splatting~\cite{KerblKLD2023} & 23.76 & 0.830 & 0.248 & \textbf{94.95} \\
		VR-NeRF~\cite{XuALGBKRPKBLZR2023} & 24.00 & 0.814 & 0.301 & 5.38 \\
		Zip-NeRF~\cite{BarroMVSH2023} & \textbf{24.79} & \textbf{0.863} & \textbf{0.216} & \textless 0.1 \\
            \midrule
		\method & \underline{24.64} & \underline{0.835} & \underline{0.236} & 41.90 \\
		
		\bottomrule
		\end{tabular}%
	}
     \\[3pt]{\footnotesize * Our implementation. VolSDF: with iNGP acceleration.}
\label{table:snpp-results}
\vspace*{-4mm}
\end{table}

\subsection{Diagnostics}
\label{sec:diagnostics}

\paragraph{Methods}

We ablate our design decisions by individually omitting the major components of our method, most notably: our distance-adjusted Eikonal loss, our adaptive surfaceness
$\beta(\vct{x})$, MLP distillation, and hardware-accelerated textures (vs. iNGP \cite{MuelleESK2022} hash tables commonly used by other fast NeRF methods).

\paragraph{Results}

We present results against the Eyeful Tower \cite{XuALGBKRPKBLZR2023} in \cref{table:diagnostics}.
Spatially adaptive surfaceness is crucial as using a global parameter degrades either speed (when $\beta$ is optimized for quality) or rendering quality (when set for speed).
Applying uniform Eikonal loss instead of our distance-adjusted variant degrades quality in unbounded scenes.
Omitting the distillation process has a minor impact on quality relative to rendering speed.
We note a similar finding when using iNGP \cite{MuelleESK2022} primitives instead of CUDA textures, which suggests that introducing hardware acceleration into these widely used primitives is a potential avenue for future research.

\section{Limitations}

\paragraph{Memory}

Storing features in dense 3D grids and triplanes consumes significantly more memory than with hash tables \cite{MuelleESK2022}.
Training is especially memory-intensive as intermediate activations must be stored for backpropagation along with per-parameter optimizer statistics.
Storing features in a hash table during the training phase before ``baking" them into explicit textures as in MERF \cite{ReiseSVSMGBH2023} would ameliorate training-time consumption but not at inference time.

\paragraph{Training time.}

Although our training time is much faster than the original NeRF, it is about 2$\times$ slower than iNGP due to the additional backprogation needed for Eikonal regularization (in line with other ``fast" surface approaches such as NeuS-facto \cite{Yu2022SDFStudio}), and slower than 3D Gaussian splatting.

\section{Conclusion}

We present a hybrid surface--volume representation that combines the best of surface and volume-based rendering into a single model.
We achieve state-of-the-art quality across several datasets while maintaining real-time frame rates at VR resolutions.
Although we push the performance frontier of raymarching approaches, a significant speed gap remains next to splatting-based approaches \cite{KerblKLD2023}.
Combining the advantages of our surface--volume representation with these methods would be a valuable next step.

{
    \small
    \bibliographystyle{ieeenat_fullname}
    \bibliography{main}
}

\clearpage

\maketitlesupplementary
\appendix

\section{Color Distillation}

\begin{figure}
    \centering
    \includegraphics[width=\linewidth, clip, trim = 0mm 2mm 0mm 0mm]{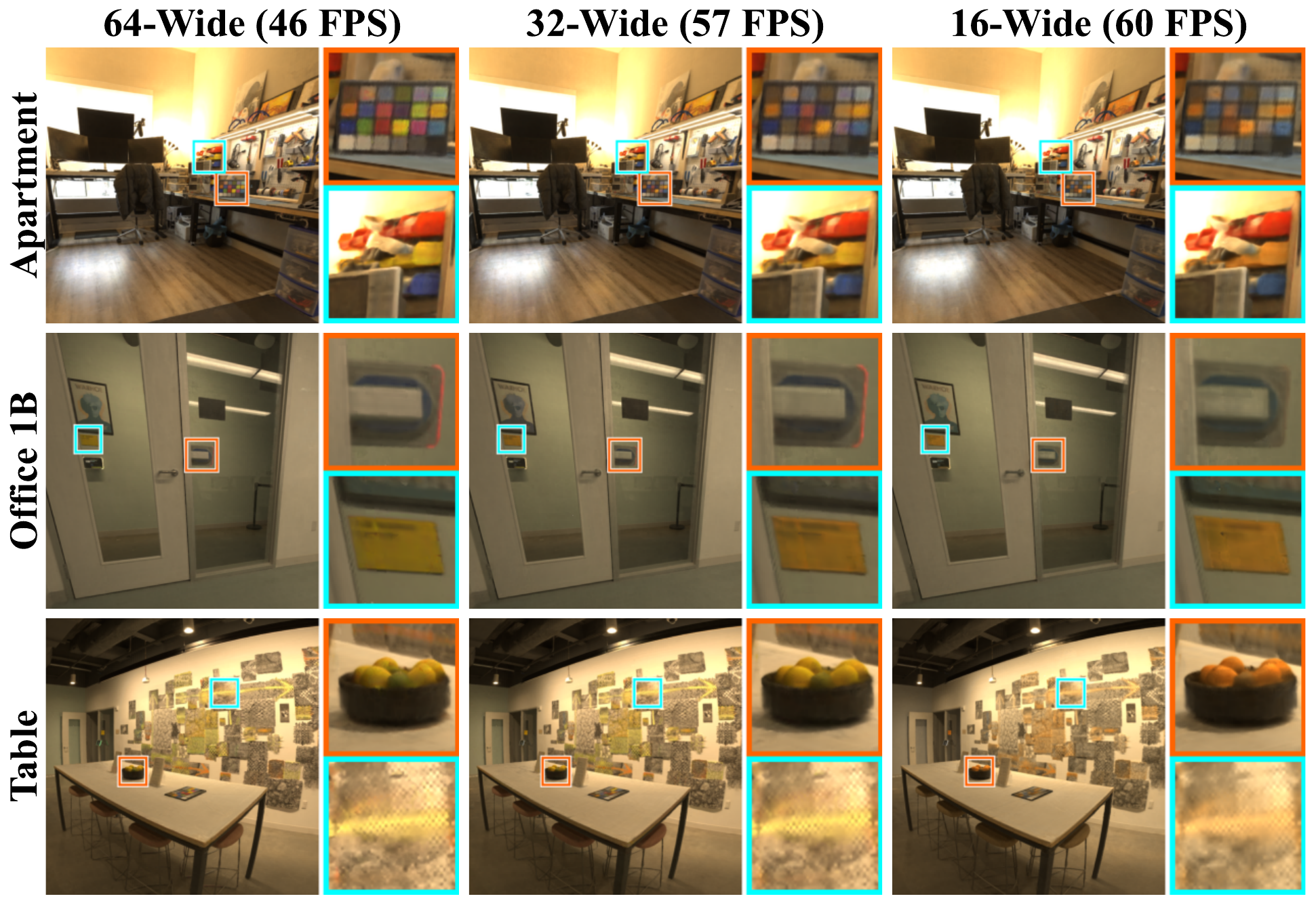}
    \caption{\textbf{Color Distillation.}
        Distilling the color MLP to a smaller width during the finetuning stage (\cref{sec:finetuning}) accelerates rendering at the cost of a minor decrease in quality.
        We observe largely similar results when decreasing the width to 32 channels, and more noticeable changes in color when further decreasing to 16.}
    \label{fig:distilled-color-qualitative}
    \vspace*{-4mm}

\end{figure}

We distill the MLP used to represent distance from our 256-wide MLP to a 16-wide network during the finetuning stage (\cref{sec:finetuning}).
It is possible to further accelerate rendering by similarly distilling the color MLP.
We found this to provide a significant boost in rendering speed (from 46 to 60 FPS) at the cost of a minor but statistically significant decrease in rendering quality (see \cref{table:distilled-color}).
We observed qualitatively similar results when decreasing width from 64 to 32 channels with more notable changes in color when decreasing the width to 16 channels (see \cref{fig:distilled-color-qualitative}).
As our initial results suggest that MLP evaluation remains a significant rendering bottleneck, replacing our scene-wide color MLP with a collection of smaller, location-specific MLPs, as suggested by KiloNeRF \cite{ReisePLG2021} and SMERF \cite{DuckwHRZTLSB2023}, is potential future work that could boost rendering speed at a smaller cost in quality.

\section{Anti-Aliasing}
\label{sec:lod}

We model rays as cones \cite{BarroMTHMS2021} and use a similar anti-aliasing strategy to VR-NeRF \cite{XuALGBKRPKBLZR2023} by dampening high-resolution grid features based on pixel footprint.
For a given sample $\vct{x}$, we derive a pixel radius $p(\vct{x})$ in the contracted space, and calculate the optimal feature level $L(\vct{x})$ based on the Nyquist--Shannon sampling theorem:
\begin{equation}
\label{eq:dampening-level}
    L(\vct{x}) \coloneqq -\log_2(2s \cdot p(\vct{x})) \text{,}
\end{equation}
where $s$ is our base grid resolution (128).
We then multiply grid features at resolution level $L$ with per-level weights $w_L$:
\vspace{-\baselineskip} 
\begin{align}
\label{eq:dampening-weight}
    w_L = \begin{cases}
        1 & \text{if } L < \floor{L(\vct{x})} \\
	L(\vct{x}) - \floor{L(\vct{x})} & \text{if } \floor{L(\vct{x})} < L \leq \ceil{L(\vct{x})} \\
        0 & \text{if } \ceil{L(\vct{x})} < L \text{.}
    \end{cases}
\end{align}

\section{Model architecture}
\label{sec:model-architecture}

We render color and distance as follows:
\begin{align}
  \vct{c}(\vct{x}, \vct{d}) &= \text{MLP}_\text{col}(\Gamma_\text{col}(\vct{x}), \text{SH}(\vct{d})) \label{eq:color-mlp} \\
  f(\vct{x}) &= \text{MLP}_\text{dist}(\Gamma_\text{dist}(\vct{x})) \text{,}
\end{align}
where $\Gamma_\text{col}$ and $\Gamma_\text{dist}$ are separate spatial feature encodings:
\vspace{-\baselineskip} 
\begin{align}
	\Gamma_\bullet(\vct{x}) &= \bigoplus_{g \in \{G_\bullet, T_\bullet^1, T_\bullet^2, T_\bullet^3\}} \sum_{l=0}^{L_g-1} w_l \cdot g(\vct{x}, l) \text{.}
	\label{eq:res-features}
\end{align}
Here, $L_g$ is the number of levels in the 3D grid $G_\bullet$ and triplanes $\{T_\bullet^1, T_\bullet^2, T_\bullet^3\}$,
$g(\vct{x}, l)$ is the (interpolated) feature vector at $\vct{x}$ for level $l$,
$w_l$ is a per-level dampening weight for anti-aliasing (\cref{eq:dampening-weight}) and `$\bigoplus$' is concatenation.
We encode the direction $\vct{d}$ through spherical harmonics, $\text{SH}(\vct{d})$, as an auxiliary input to the color MLP (\cref{eq:color-mlp}) that is independent of the feature vector $\Gamma_\text{col}(\vct{x})$.

\begin{table}
    \caption{\textbf{Color distillation.}
    We evaluate the effect of color MLP distillation on the Eyeful Tower dataset \cite{XuALGBKRPKBLZR2023}, and find a significant increase in rendering speed at the cost of quality.}
	\centering
	\resizebox{\linewidth}{!}{%
		\begin{tabular}{l@{\hspace{1em}}c@{\hspace{1em}}c@{\hspace{1em}}c@{\hspace{1em}}r}
		\toprule 
		Color Width & $\uparrow$PSNR & $\uparrow$SSIM & $\downarrow$LPIPS & $\uparrow$FPS \\ \midrule
		16-wide (distilled) & 30.88 & 0.888 & 0.236 & \textbf{60.13} \\
		32-wide (distilled) & \underline{31.17} & \underline{0.900} & \underline{0.220} & \underline{57.05} \\
		64-wide (original) & \textbf{31.57} & \textbf{0.913} & \textbf{0.198} & 45.78 \\
		\bottomrule
		\end{tabular}%
	}
\label{table:distilled-color}
\vspace*{-4mm}

\end{table}

\begin{table*}
    \caption{\textbf{Grid feature layout.}
    We measure the effect of using only 3D or triplane features on the Eyeful Tower dataset \cite{XuALGBKRPKBLZR2023}, and note a significant drop in quality when compared to using both.}
	\centering
		\begin{tabular}{l@{\hspace{1em}}c@{\hspace{1em}}c@{\hspace{1em}}c@{\hspace{2em}}c@{\hspace{1em}}c@{\hspace{1em}}c@{\hspace{2em}}c@{\hspace{1em}}c@{\hspace{1em}}r}
		\toprule
		& \multicolumn{3}{c@{\hspace{2em}}}{Pinhole}
		& \multicolumn{3}{c@{\hspace{2em}}}{Fisheye}
		& \multicolumn{3}{c}{Overall}
		\\ \cmidrule(r{2em}){2-4}\cmidrule(r{2em}){5-7}\cmidrule(r{0.6em}){8-10}
		& $\uparrow$PSNR & $\uparrow$SSIM & $\downarrow$LPIPS
		& $\uparrow$PSNR & $\uparrow$SSIM & $\downarrow$LPIPS
		& $\uparrow$PSNR & $\uparrow$SSIM & $\downarrow$LPIPS \\ \midrule
		3D Only & 27.10 & 0.832 & 0.410	& 32.17 & 0.928 & 0.187 & 29.41 & 0.875 & 0.308 \\
		Triplane Only & \underline{28.24} & \underline{0.843} & \underline{0.312} & \underline{33.16} & \underline{0.938} & \underline{0.150} & \underline{30.47} & \underline{0.886} & \underline{0.238} \\
  \midrule
		Both & \textbf{29.07} & \textbf{0.880} & \textbf{0.268} & \textbf{34.57} & \textbf{0.952} & \textbf{0.115} & \textbf{31.57} & \textbf{0.913} & \textbf{0.198} \\
		\end{tabular}%
\vspace*{-4mm}
\label{table:grid-layout}
\end{table*}

We use low-resolution 3D grids and high-resolution triplanes, as in previous work (MERF \cite{ReiseSVSMGBH2023}) to obtain the best rendering quality (\cref{table:grid-layout}).
We double the grid resolution between levels and therefore have a low-resolution 3D grid with 3 levels ($128^3$–$512^3$) and higher-resolution triplanes with 7 levels ($128^3$–$8192^3$).
Na\"ively computing \cref{eq:res-features} requires $3 \!+\! 3 \!\times\! 7 = 24$ \textit{texture} fetches (we rely on the CUDA texture API for hardware interpolation and do not need to explicitly query voxel corners/texels).
As a render-time optimization, we save pre-summed features $g'_L$ for each resolution level $L$ (where we store $g'_L(\vct{v}) = \sum_{l=0}^{L} g(\vct{v}, l)$ for each texel $\vct{v}$ in $L$), such that~\cref{eq:res-features} can be rewritten as $\Gamma(\vct{x}) = \bigoplus_{g \in \{G,T^1,T^2,T^3\}} [w_{L}g'_{L}(\vct{x}) + (1 - w_{L})g'_{L - 1}(\vct{x})]$ for $L \!=\! L(\vct{x})$ (\cref{eq:dampening-level}).
Here, $\vct{v}$ refers to the texel (voxel).
Querying two levels requires only $2 + 3 \times 2 = 8$ texture fetches.
\looseness-1

\section{Geometric Reconstruction}

We evaluate geometric reconstruction on ScanNet++ \cite{YeshwLND2023} (which has ``ground-truth" laser scan depth only for foreground pixels) in \cref{table:depth-results} for the strategies in \cref{fig:eikonal-scaling}.
Using uniform Eikonal loss in contracted space degrades accuracy (0.419\,m error vs 0.219\,m for uniform world space and 0.221\,m with our distance-adjusted method) and omitting Eikonal loss gives the worst results (0.996\,m).

\section{ScanNet++}

We evaluate 9 scenes from ScanNet++ \cite{YeshwLND2023} in \cref{sec:common-eval} ({\footnotesize\textsc{5fb5d2dbf2, 8b5caf3398, 39f36da05b, 41b00feddb, 56a0ec536c, 98b4ec142f, b20a261fdf, f8f12e4e6b, fe1733741f})}.
We undistort the fisheye DSLR captures to pinhole images using the official dataset toolkit \cite{snpp_toolkit} to facilitate comparisons against 3D Gaussian splatting \cite{KerblKLD2023} (whose implementation does not support fisheye projection).
We use the official validation splits, which consist of entirely novel trajectories that present a more challenging novel-view synthesis problem than the commonly used pattern of holding out every eighth frame \cite{MildeSOKRNK2019}.
The dataset authors note that their release is still in the beta testing phase, and that the final layout is subject to change.
Our testing reflects the dataset as of November 2023.

\begin{table}
    \caption{\textbf{Depth error on ScanNet++ \cite{YeshwLND2023}.} Our distance-adjusted Eikonal loss degrades geometric reconstruction less than other alternatives used to render unbounded scenes.}
	\centering
	\resizebox{\linewidth}{!}{%
		\begin{tabular}{l@{\hspace{1em}}c@{\hspace{1em}}r}
		\toprule 
		Method & $\downarrow$Distance (m) & $\downarrow$Distance (\%) \\ \midrule
		Uniform Eikonal loss (world space) & \textbf{0.219} & \textbf{8.56}  \\
            Uniform Eikonal loss (contracted space) & 0.419 & 16.11  \\
		No Eikonal loss & 0.996	& 29.93  \\
		\midrule
		Distance-adjusted Eikonal loss (ours) & \underline{0.221} & \underline{11.13}  \\
		\bottomrule
		\end{tabular}%
	}
\label{table:depth-results}
\vspace*{-4mm}
\end{table}

\section{Societal Impact}

Our technique facilitates the rapid generation of high-quality neural representations.
Consequently, the risks associated with our work parallel those found in other neural rendering studies, primarily centered around privacy and security issues linked to the deliberate or unintentional capture of sensitive information, such as human facial features and vehicle license plate numbers.
Although we did not specifically apply our approach to data involving privacy or security concerns, there exists a risk, akin to other neural rendering methodologies, that such sensitive data could become incorporated into the trained model if the datasets utilized are not adequately filtered beforehand.
It is imperative to engage in pre-processing of the input data employed for model training, especially when extending its application beyond research, to ensure the model's resilience against privacy issues and potential misuse.
However, a more in-depth exploration of this matter is beyond the scope of this paper.

\end{document}